\documentclass[runningheads]{llncs}

\usepackage{eccv}

\usepackage{eccvabbrv}

\usepackage{graphicx}
\usepackage{booktabs}

\usepackage[accsupp]{axessibility}  

\usepackage[pagebackref,breaklinks,colorlinks,citecolor=eccvblue]{hyperref}

\usepackage{orcidlink}
\usepackage{adjustbox} 
\usepackage[dvipsnames]{xcolor}

\usepackage{booktabs} 

\usepackage[noend]{algpseudocode} 
\usepackage[ruled,vlined]{algorithm2e}
\usepackage{nicefrac}
\usepackage{bm}
\usepackage{multirow}
\usepackage{balance}

\usepackage{mdframed}
\definecolor{theoremcolor}{rgb}{0.94, 0.94, 0.94}
\definecolor{examplecolor}{rgb}{1, 1, 1.0}
\mdfsetup{
    backgroundcolor=theoremcolor,
    linewidth=0pt,
}
\usepackage{xfrac}
\usepackage{comment}
\newmdtheoremenv[linewidth=0pt,innerleftmargin=4pt,innerrightmargin=4pt]{prop}{Proposition}
\newmdtheoremenv[linewidth=0pt,innerleftmargin=4pt,innerrightmargin=4pt]{assump}{Assumption}
\newmdtheoremenv[linewidth=0pt,innerleftmargin=4pt,innerrightmargin=4pt]{defn}{Definition}
\newcommand{\E}{\mathbb{E}}
\newcommand{\R}{\mathbb{R}}

\newcommand{\X}{\mathcal{X}}

\newcommand{\gL}{\mathcal{L}}
\newcommand{\Z}{\mathcal{Z}}

\def\mA{{\mathbf{A}}}

\def\mI{{\mathbf{I}}}

\def\rvx{{\mathbf{x}}}
\def\rvy{{\mathbf{y}}}
\def\rvz{{\mathbf{z}}}

\usepackage{mathtools}
\DeclarePairedDelimiterX{\infodivx}[2]{(}{)}{%
	#1\;\delimsize\|\;#2%
}
\newcommand{\KL}{D_\mathrm{KL}}
\DeclareMathOperator*{\argmin}{arg\,min}

\newcommand{\dec}{\beta}
\newcommand{\enc}{\alpha}
\newcommand{\energy}{\omega}
\newcommand{\prior}{\theta}
\newcommand{\energyfn}{E_\omega}
\newcommand{\encoder}{f_{\enc}}
\newcommand{\decoder}{g_{\dec}}

\newcommand{\name}{Energy-Calibrated VAE\xspace}
\newcommand{\shortname}{EC-VAE\xspace}
\newcommand{\spara}[1]{\vspace{1mm}\noindent\textbf{#1.}}

\begin{document}

\title{\name with Test Time Free Lunch} 

\titlerunning{Energy-Clibrated VAE}

\author{
Yihong Luo\inst{1,2} \and
Siya Qiu\inst{1,2} \and 
Xingjian Tao\inst{2} \and
Yujun Cai\inst{3} \and
Jing Tang\inst{2, 1}\thanks{Corresponding author: Jing Tang.}
}

\authorrunning{Y. Luo et al.}

\institute{$^\text{1 }$The Hong Kong University of Science and Technology  \\
$^\text{2 }$The Hong Kong University of Science and Technology (Guangzhou)
~~~$^\text{3 }$Meta \\
\email{yluocg@connect.ust.hk, jingtang@ust.hk}\\
}

\maketitle

\begin{abstract}
In this paper, we propose a novel generative model that utilizes a conditional Energy-Based Model (EBM) for enhancing Variational Autoencoder (VAE), termed Energy-Calibrated VAE (EC-VAE). Specifically, VAEs often suffer from blurry generated samples due to the lack of a tailored training on the samples generated in the generative direction. On the other hand, EBMs can generate high-quality samples but require expensive Markov Chain Monte Carlo (MCMC) sampling. To address these issues, we introduce a conditional EBM for calibrating the generative direction of VAE during training, without requiring it for the generation at test time. In particular, we train EC-VAE upon both the input data and the calibrated samples with adaptive weight to enhance efficacy while avoiding MCMC sampling at test time.
Furthermore, we extend the calibration idea of EC-VAE to variational learning and normalizing flows, and apply EC-VAE to an additional application of zero-shot image restoration via neural transport prior and range-null theory.
We evaluate the proposed method with two applications, including image generation and zero-shot image restoration, and the experimental results show that our method achieves competitive performance over single-step non-adversarial generation.
\end{abstract}
\vspace{-0.5mm}
\section{Introduction}
\label{sec:intro}
Deep generative models, including Generative Adversarial Nets (GANs)~\cite{goodfellow2014generative}, Variational Autoencoders (VAEs)~\cite{kingma2014vae}, flow-based generative models~\cite{niceflow,dinh2016density}, Energy-Based Models (EBMs)~\cite{du2019implicit,xie2016theory}, and diffusion models~\cite{ddpm}, achieve excellent performance in a variety of applications. Compared to GANs, likelihood-based models, such as VAEs and EBMs, typically exhibit greater stability during training and more faithfully cover modes in the data. Moreover, unlike normalizing flows that suffer from architecture restrictions~\cite{kong2020expressive}, VAEs and EBMs offer considerable potential for expressivity. Therefore, VAEs and EBMs have gained extensive attention recently.

In particular, VAEs map the input data into a latent distribution and optimize the evidence lower bound (ELBO) on the data likelihood. 
However, VAEs do not explicitly optimize the generative direction. Specifically, VAEs assume that the prior distribution in latent space (e.g., Gaussian distribution) matches the empirical distribution mapped from the input data. Unfortunately, there is often a gap between the two distributions in practice. As a result, VAEs struggle to generate high-quality images, producing blurry or corrupted samples. 
To tackle this issue, in addition to designing more flexible prior distribution~\cite{vamprior,yang2019pointflow,nvp-vae}, some work~\cite{tzeng2017adversarial,dcvae, huang2018introvae} involves GANs, yielding unstable adversarial games, while NVAE~\cite{vahdat2020NVAE} proposes to adjust BatchNorm~\cite{batchnorm} statistics based on prior distribution that improves the generation slightly. 

On the other hand, EBMs directly model the unnormalized density in data space by assigning low energy to high-probability areas. 
Unlike VAEs that usually assume a Gaussian prior, EBMs do not necessitate distribution assumptions in modeling.
In fact, EBMs have shown comparable state-of-the-art performance in terms of generative results among non-adversarial methods~\cite{clel}.
However, a significant drawback of EBMs is the necessity for Markov Chain Monte Carlo (MCMC) sampling during the training and during the generation at test time, which suffers from slow convergence and is computationally expensive, particularly when the energy is parameterized by neural networks.

In this paper, we propose a novel generative model termed \name (\shortname) by involving a conditional EBM to calibrate the VAE for better generation while keeping high sampling efficiency. The VAE is trained by ELBO and the energy-based calibration.
Specifically, to address the training of the generative side (\ie, the generating process from prior to data) that is missing in the conventional VAE, we propose to incorporate the generated samples into training.
That is, for the generated data $\hat{\rvx}$ by the generative model, we use a conditional EBM to sample data $\Tilde{\rvx}$ initialized with $\hat{\rvx}$ that approximates the input real data. 
Then, integrating the minimization of the distance between $\hat{\rvx}$ and $\Tilde{\rvx}$ into training can calibrate the decoder.
We show that the proposed model can be jointly trained by adopting the primal-dual method in a constrained formulation. 
The conditional EBM is solely utilized for calibrating samples. Consequently, a short-run MCMC sampling is sufficient and does not suffer from slow convergence.
Note that the EBM is involved in training (i.e.,~generation calibration) only, and is not required during test time sampling.

Moreover, we show that the idea of energy-based calibration can be extended to calibrate the variational inference and normalizing flows with large improvement. Take the former as an illustration while the latter is similar. We first sample $\rvz$ from variational posterior, and then calibrate $\rvz$ by running MCMC sampling on constructed conditional posterior $p(\Tilde{\rvz}|\rvz,\rvx)$ to obtain the calibrated $\Tilde{\rvz}$. Finally, minimizing the distance between $\rvz$ and $\Tilde{\rvz}$ can calibrate the encoder. 

Experimental results show that the proposed \shortname outperforms previous EBMs and the state-of-the-art VAEs on image generation benchmarks in both low-resolution and high-resolution datasets by a large margin. In particular, \shortname achieves the strong performance with a single-step non-adversarial generation manner, competing with advanced GANs and diffusions across multiple datasets.
We also propose and show how to apply \shortname to image restoration in a zero-shot way by constructing neural transport prior and leveraging range-null space theory with competitive performance.


Our main contributions are summarized as follows.
\begin{enumerate}
    \setlength{\topsep}{-5px}
    \setlength{\itemsep}{0pt}
    \setlength{\partopsep}{0pt}
    \setlength{\parskip}{0pt}
    \setlength{\parsep}{0pt}
    \item We propose a new generative model termed \shortname utilizing a conditional EBM to calibrate the VAE to generate sharper samples without incurring extra costs of MCMC sampling during the generation at test time. 
    \item We extend the energy-based calibration to enhance variational learning and normalizing flows, and apply \shortname to an additional application of zero-shot image restoration.
    \item We demonstrate the strong empirical results of our proposed methods on various tasks, including image generation and image restoration. 
\end{enumerate}

\vspace{-2mm}
\section{Preliminaries}
\spara{Notations} 
Denote by $\rvx$ the data and by $\rvz$ the latent variable. Let $\X$ be the data space and $\Z$ be the latent space. Let $p_d(\rvx)$ be the data distribution. 
Denote by $\encoder\colon \thickmuskip=2mu \medmuskip=2mu \X \rightarrow \Z$ the encoder parameterized by $\enc$, and by $\decoder\colon \thickmuskip=2mu \medmuskip=2mu \Z\rightarrow \X$ the decoder parameterized by $\dec$. 
Denote by $\energyfn\colon \thickmuskip=2mu \medmuskip=2mu \X \rightarrow \R$ the energy function parameterized by $\energy$. 

\spara{Variational Autoencoders}
VAE~\cite{kingma2014stochastic} adopts the encoder-decoder architecture with a prior distribution. To be more precise, VAE defines the joint distribution of $(\rvx,\rvz)$ as $p_{\dec,\prior}(\rvx,\rvz) = p_\dec(\rvx|\rvz)p_\prior(\rvz)$. The model can be trained by maximizing the marginal log-likelihood $\gL = \E_{p_d(\rvx)}[\log p_{\dec,\prior}(\rvx)]$. However, maximizing the log-likelihood needs sampling from intractable posterior $p_{\dec,\prior}(\rvz|\rvx) = \frac{p_{\dec,\prior}(\rvx,\rvz)}{p_{\dec,\prior}(\rvx)}$.

Instead of sampling from intractable posterior $p_{\dec,\prior}(\rvz|\rvx)$, VAEs propose using a variational inference $q_\enc(\rvz|\rvx)$ to approximate the posterior. In particular, VAEs optimize the evidence lower bound (ELBO) on $\log p_{\dec,\prior}(\rvx)$ such that
\begin{equation}
\small
\mathrm{ELBO}_{\phi}(\rvx) = \E_{q_\enc(\rvz|\rvx)}\big[\log p_\dec(\rvx|\rvz))-\mathrm{KL}(q_\enc(\rvz|\rvx)||p_\prior(\rvz)\big],
\end{equation}
where $\phi = \{ \enc,\dec,\prior \}$. The first term is the reconstruction loss and the second term is the KL divergence between the approximated posterior and the prior.

Sampling from VAE can be achieved by sampling $\rvz$ from prior $p_\prior(\rvz)$ first, and then obtain generated samples $\rvx$ with probability $p_\dec(\rvx|\rvz)=\mathcal{N}(g_\dec(\rvz),\sigma^2\mathbf{I})$. Equivalently, we denote samples $\rvx\sim p_{\dec,\prior}(\rvx)= \int p_\prior(\rvz)p_{\dec,\prior}(\rvx|\rvz)\mathrm{d}\rvz$. In practice, the samples are typically obtained directly by $g_\dec(\rvz)$.

\spara{Energy-Based Models}
A deep EBM assumes that $p_\energy(\rvx)$ is a gibbs distribution with the form $p_\energy(\rvx) = \exp(-E_\energy(\rvx))/L_\energy$, where $L_\energy$ is the corresponding normalizing constant. EBM is trained by minimizing the negative log-likelihood~(NLL) $\gL(\energy)$ such that 
\begin{equation*}
\small
    \gL(\energy) = -\E_{p_d(\rvx)} [\log p_\energy(\rvx)] = -\E_{p_d(\rvx)}\big[\tfrac{\exp(-E_\energy(\rvx))}{L_\energy}\big].
\end{equation*}
The gradient of $\gL(\energy)$ can be obtained as follows:
\begin{equation}
\small
\label{eq:ebm_grad}
\nabla \gL(\energy)  = \E_{p_d(\rvx)} [\nabla E_\energy(\rvx)] - \E_{p_\energy(\rvx)} [\nabla E_\energy(\rvx)].
\end{equation}
In practice, sampling from $p_\energy(\rvx)$ can be achieved by running MCMC sampling with $K$ steps of Langevin dynamics, with initial samples $\rvx_0$ and step size $s$:
\begin{equation}
\small
\notag
    \rvx_{k+1} = \rvx_{k} - \frac{s}{2}\nabla_{\rvx_k} E_\energy(\rvx_{k}) + \sqrt{s} \xi, \quad\text{where } \xi \sim \mathcal{N}(0,\mI).
    \label{eq: LD}
\end{equation}

\vspace{-6mm}
\subsection{Additional Related Work}
\label{sec:related_work}
Our method shares some similarities with works that combine generative autoencoders and EBMs in different ways. VAEBM~\cite{vaebm} learns EBM upon a pre-trained NVAE~\cite{vahdat2020NVAE}, and a recent work~\cite{han2020joint} jointly learns EBM and VAE by an adversarial game instead of by MCMC sampling, while our work is jointly trained without adversarial components. 
In addition, our learning algorithm bears some similarities to cooperative learning~\cite{xie2018cooperative,coopvaebm,coopflow,hatebm}, which also employs EBM to teach the base generative model, but only in small images. 
However, the base generative model in these approaches is purely trained upon \textit{generated samples}, which can be quite biased. 
In contrast, our base generative model (i.e., VAE) is trained upon data and generated samples, with adaptive weights. 
Recently, dual MCMC~\cite{dualmcmc} also consider incorporating real data into cooperative-like training. However, similar to other cooperative approaches~\cite{xie2018cooperative,coopvaebm}, they maximize the \textit{marginal likelihood} of the generated samples by latent variable models. This requires extra effort in inferring latent variables, increasing the burden of the model and training cost. In contrast, we maximize the \textit{conditional likelihood} of the generated samples via the decoder of VAE, which is easier to learn. 
Moreover, our method differs significantly from aforementioned works in the way that we discard MCMC during inference without adversarial components, providing a strong one-step generation that competes with diffusions and GANs.

\vspace{-1mm}
\section{The Design of \name}
\label{sec:energy calibration}
An issue with VAE is that the generative direction has not been explicitly trained during the training process, potentially leading to lower-quality output for the generated samples. To address this, we propose to incorporate the generated samples into the training. As the real data corresponding to the generated samples are unavailable, we propose utilizing a short-run MCMC, initialized with the generated samples, to approximate the corresponding real data.

As the aim is to calibrate the samples, we suggest constructing a \textbf{conditional EBM}, similar in \cite{gao2020learning}, to model the conditional density, i.e.,
\begin{equation}
\small
\notag
    p_\energy(\Tilde{\rvx}|\rvx) = \frac{1}{L_\energy(\rvx)} \mathrm{exp}(-\energyfn(\Tilde{\rvx}) - \frac{||\Tilde{\rvx}-\rvx ||_2^2}{2\sigma^2} ),
\end{equation}
where $\sigma$ is a pre-defined hyper-parameter, $L_\energy(\rvx) = \int \mathrm{exp}(-\energyfn(\Tilde{\rvx}) - \frac{||\Tilde{\rvx}-\rvx ||_2^2}{2\sigma^2})\mathrm{d}\Tilde{\rvx}$ is the corresponding normalizing constant, $\rvx$ is the generated samples from VAE, $\energyfn$ is an unconditional EBM. Compared to direct model $p_\energy(\Tilde{\rvx})$, the extra distance term in $p_\energy(\Tilde{\rvx}|\rvx)$ constrains the high-density area localized around generated samples $\rvx$, making it easier to be learned by the base generative model (i.e., VAE). 
Sampling from $p_\energy(\Tilde{\rvx}|\rvx)$ can be achieved by MCMC sampling, i.e.,
\begin{equation}\label{eq: LD_cond}
\small
    \rvx_{k+1} = \rvx_{k} - \frac{s}{2}\Big( \underbrace{\nabla_{\rvx_k} E_\energy(\rvx_{k})}_\text{direct to real} + \underbrace{\nabla_{\rvx_k} \frac{||\rvx_k - \rvx ||_2^2}{2\sigma^2}}_\text{direct to origin} \Big) +  \sqrt{s} \xi, \quad\text{where } \xi \sim \mathcal{N}(0,\mI).
\end{equation}
For simplicity, we denote $\rvx^K = \operatorname{MCMC}_\energy^K(\Tilde{\rvx}|\rvx)$.  
The learning of EBM can be achieved by minimizing $\gL(\energy) \triangleq \E_{p_{\dec,\prior}(\rvx)}[\KL(p_d(\Tilde{\rvx}|\rvx) || p_\energy(\Tilde{\rvx}|\rvx) )]$: 
\begin{equation*}
\vspace{-1mm}
\small
\begin{split}
    \nabla \gL(\energy) 
    & = \E_{p_{\dec,\prior}(\rvx)} \big[
        \E_{p_d(\Tilde{\rvx}|\rvx)} [\nabla_\energy \log p_\omega(\Tilde{\rvx}|\rvx)]
        - \E_{p_\energy(\Tilde{\rvx}|\rvx)} [\nabla_\energy \log p_\omega(\Tilde{\rvx}|\rvx)]
    \big] \\
    &= \E_{p_{\dec,\prior}(\rvx)}[\E_{p_d(\Tilde{\rvx}|\rvx)} [\nabla_\energy \energyfn(\Tilde{\rvx})]] - \E_{p_{\dec,\prior}(\rvx)}[\E_{p_\energy(\Tilde{\rvx}|\rvx)}[ \nabla_\energy \energyfn(\Tilde{\rvx})]] \\
    &= \E_{p_d(\Tilde{\rvx})} [\nabla_\energy \energyfn(\Tilde{\rvx})] - \E_{p_{\dec,\prior}(\rvx)}[\E_{p_\energy(\Tilde{\rvx}|\rvx)} [\nabla_\energy \energyfn(\Tilde{\rvx})]]. \\
\end{split}
\vspace{-2mm}
\end{equation*}
Since the distance term does not involve learnable parameters, hence the parameter $\energy$ can be learned in unconditional form. 
We only need to define $p_d(\Tilde{\rvx}|\rvx)$ to ensure the marginal distribution of $p_{\dec,\prior}(\rvx)p_d(\Tilde{\rvx}|\rvx)$ being data distribution (e.g., the simplest case is $p_d(\Tilde{\rvx}|\rvx) \triangleq p_d(\Tilde{\rvx})$).  It can be seen that the $\gL(\energy)$ reaches minima at $\int p_{\dec,\prior}(\rvx)p_\energy(\Tilde{\rvx}|\rvx)d\rvx = p_d(\Tilde{\rvx})$. This is said that given rich enough $p_\energy(\Tilde{\rvx}|\rvx)$, we are able to calibrate samples from $p_{\dec,\prior}(\rvx)$ to $p_d(\Tilde{\rvx})$. 

Then, we regard the $\Tilde{\rvx}$ as calibrated samples, thus the generative direction can be calibrated by minimizing the distance between $\rvx$ and $\Tilde{\rvx}$, 
\begin{equation*}
\vspace{-1mm}
\small
\begin{split}
    \gL_{\mathrm{calibration}} 
    &= \E_{p_{\dec,\prior}(\rvx)} [\E_{p_\energy(\Tilde{\rvx}|\rvx)} [||\rvx - \Tilde{\rvx}||_2^2]].
\end{split}
\end{equation*}
Note that the calibration loss can be considered as maximizing the likelihood of calibrated samples conditioned on generated samples under the form of normal distribution.
This MCMC for sampling calibrated $\Tilde{\rvx}$ can be regarded as a teacher, guiding the generative model to produce higher-quality generated samples. However MCMC sampling introduces some random noise into training, making it impossible to perfectly match the calibrated samples, thus we suggest using a constrained optimization form as follows:
\vspace{-1.mm}
\begin{defn}[\name]\label{defn: enhanced constrained opt}
Given a fixed margin $\epsilon_1$, the general optimization  can be transformed into the following
inequality-constrained optimization.
\vspace{-1.5mm}
\begin{equation*}
\small
    \begin{split}
    \min_{\phi,\energy} \quad &\gL(\phi) + \gL(\energy)\\
    \mathrm{s.t.}\quad & ||\Tilde{\rvx}-\rvx ||_2^2 < \epsilon_1,\ \forall \rvx\sim p_{\dec,\prior}(\rvx), \Tilde{\rvx} = \operatorname{MCMC}^K_\energy(\Tilde{\rvx}|\rvx),\\
    \mathrm{where} \ \ & \small \gL(\energy) = \E_{p_{\dec,\prior}(\rvx)}\KL(p_d(\Tilde{\rvx}|\rvx) || p_\energy(\Tilde{\rvx}|\rvx) ), \\
    & \small \gL(\phi) = -\E_{p_d(\rvx)}\E_{q_\enc(\rvz|\rvx)}\big[\log p_\dec(\rvx|\rvz) - \tfrac{q_\enc(\rvz|\rvx)}{p_\prior(\rvz)}\big].\\
    \end{split}
\vspace{-1mm}
\end{equation*}
\end{defn}

We get a constrained optimization problem in Definition~\ref{defn: enhanced constrained opt}, which is hard to optimize directly, but we can consider it as a corresponding saddle-point problem as follows:
\vspace{-2mm}
\begin{equation}
\small
\notag
    \max\limits_{\lambda} \min\limits_{\phi,\energy} \big\{  \gL(\phi) + \gL(\energy) + 
    \lambda\gL_{\mathrm{con}}(\phi)\big\}, \quad \lambda \geq 0
\end{equation}
where the constraint-related loss $\gL_{\mathrm{con}}(\phi)$ is defined as: 
\vspace{-2mm}
\begin{equation}
\small
\notag
    \begin{aligned}
    &\gL_{\mathrm{con}}(\phi) = \gL_{\mathrm{con}}(\dec) = \E_{p_{\dec,\prior}(\rvx)}[\E_{p_\energy(\Tilde{\rvx}|\rvx)}[||\rvx-\Tilde{\rvx}||_2^2 - \epsilon_1]].
    \end{aligned}
\vspace{-1mm}
\end{equation}
Since the prior is typically less powerful, we suggest calibrating only the decoder, enabling the prior focus on maximizing likelihood related to the latent variable.

The final challenge is the lack of access to the ground truth of data distribution. To tackle this issue, we consider the corresponding empirical optimization problem as follows:
\begin{equation*}
\small
\begin{split}
    & \max\limits_{\lambda} \min\limits_{\phi,\energy} \{ \hat{\gL}(\phi) + \hat{\gL}(\energy) + \lambda\gL_{\mathrm{con}}(\phi) \}\\
    & = \max\limits_{\lambda} \min\limits_{\phi,\energy} \Big\{ 
     \sum\limits_{i=1}^{n} \sum\limits_{j=1}^{m} - \Big( \log p_{\dec}(\rvx_i|\rvz_{ij}) \mathop{+} \log\frac{q_\enc(\rvz_{ij}|\rvx_i)}{p_{\prior}(\rvz_{ij})} \Big) \mathop{+} \sum\limits_{i=1}^{n}-\log p_\energy(\rvx_i) \\
    & \phantom{=\max\limits_{\lambda} \min\limits_{\phi} \Big\{}\mathop{+} \lambda \sum\limits_{i=1}^{n} [||\hat{\rvx}_i-\operatorname{MCMC}_\energy^K(\Tilde{\rvx}_i|\hat{\rvx}_i)||_2^2 - \epsilon_1]  \Big\},
\end{split}
\end{equation*}
where $\rvx_i$ is sampled from data, $\hat{\rvx}_i$ is sampled from the VAE, $\rvz_{ij}$ is sampled from $q_\enc(\rvz|\rvx_i)$. Notice that $m$ is usually chosen to be one in practice that is efficient and effective enough to estimate the gradient. MCMC is not necessary during test time sampling.

\spara{Concrete Algorithm}
To efficiently optimize the problem, we employ the primal-dual algorithm tailored for addressing the saddle-point problem corresponding to the constrained form. Specifically, in the \textit{primal} step, the algorithm alternately optimizing parameters $\energy$ and $\phi = \{ \dec, \enc, \theta \}$ by minimizing the empirical Lagrangian under a given dual variable $\lambda$, i.e.,
\vspace{-2mm}
\begin{equation*}
\small
\begin{split}
    &\energy_{t+1}:= \argmin_{\energy} \hat{\gL}(\energy_{t}), \\
    &\dec_{t+1} := \argmin_\dec  \Big\{ \lambda \hat{\gL}_{\mathrm{con}}(\dec_{t}) - \sum\limits_{i=1}^{n} \sum\limits_{j=1}^{m}\log p_{\dec_{t}}(\rvx_i|\rvz_{ij}) \Big\}, \\
    &\enc_{t+1} := \argmin_\enc \Big\{\sum\limits_{i=1}^{n}\sum\limits_{j=1}^{m} \log\tfrac{q_{\enc_{t}}(\rvz_{ij}|\rvx_i)}{p_{\prior_{t}}(\rvz_{ij})} -\log p_{\dec_{t}}(\rvx_i|\rvz_{ij}) \Big\},\\
    &\theta_{t+1} := \argmin_\theta  \Big\{\sum\limits_{i=1}^{n}\sum\limits_{j=1}^{m}- \log p_{\prior_{t}}(\rvz_{ij})\Big\}.
\end{split}
\end{equation*}
In practice, we perform stochastic gradient descent that derives concrete update step for $\theta$ and $\phi$.
On the other hand, in the \textit{dual} step, we update $\lambda$ as follows,
\begin{equation}\label{eq:dual step}
\small
    \lambda_{t+1} := \max \left\{\lambda_{t}+\eta\cdot \big(\hat{\gL}_{\mathrm{con}}-\epsilon\big), 0\right\},
\end{equation}
where $\eta$ is the learning rate of dual step. 

\cref{algorithm:Primal-Dual} in \cref{app:algorithm} gives the pseudo-code of our primal-dual algorithm for optimizing the VAE parameters $\phi$, and EBMs parameters $\energy$. 
Compared with using stochastic gradient descent directly in the constrained optimization problem in Definition \eqref{defn: enhanced constrained opt}, the primal-dual algorithm can dynamically tune $\lambda$ to avoid introducing extra hyper-parameters (which may serve as an early-stopping condition, e.g., $\lambda=0$), and can provide convergence guarantees with sufficiently long training using sufficiently small step size~\cite{chamon2021constrained}.

\section{Extensions and Additional Application}
In this section, we first show the calibration idea of \shortname can be extended to enhance variational learning and normalizing flows. Then we show how to apply our method in zero-shot image restoration.
\subsection{Energy-Calibrated Variational Learning}
Variational learning allows efficient training, but results in learning a lower bound for data likelihood. The gap is the KL divergence between variational posterior $q_\enc(\rvz|\rvx)$ and posterior $p_{\dec,\prior}(\rvz|\rvx)$ which is not explicitly minimized in variational training, just like the training of the generative direction is missing in vanilla VAEs.

Similar to calibrating the generative direction as proposed in \cref{sec:energy calibration}, we propose to incorporate the calibration of the variational posterior $q_\enc(\rvz|\rvx)$ into training. We first construct the conditional density 
$p_{\dec,\prior}(\Tilde{\rvz}|\rvz,\rvx)$ as follows:
\begin{equation*}
        p_{\dec,\prior}(\Tilde{\rvz}|\rvz,\rvx) = p_{\dec,\prior}(\Tilde{\rvz}|\rvx) \cdot \exp(- \tfrac{||\Tilde{\rvz}-\rvz) ||_2^2}{2\sigma^2}) / L_{\dec,\prior}(\rvz),
\end{equation*}
where $\sigma$ is a pre-defined hyper-parameter, $p_{\dec,\prior}(\Tilde{\rvz}|\rvx)$ is the posterior and  $L_{\dec,\prior}(\rvz)$ is corresponding normalizing constant.
The conditional density is constructed by adding the distance term to constrain the calibrated $\rvz$ to be close to $\rvz$. And the sampling can be achieved by MCMC with Langevin dynamics. Given a step size $s>0$ and an initial value $\rvz_0$, the Langevin dynamics iterates:
\begin{equation}
\small
\notag
    \rvz_{k+1} = \rvz_{k} - \frac{s}{2}\nabla_{\rvz_k} ( -\log p_{\dec,\prior}(\rvz_k|\rvx) + \frac{||\rvz_{k}-\rvz||_2^2}{2\sigma^2} )+ \sqrt{s} \xi, \quad \xi \sim \mathcal{N}(0,\mI)
\end{equation}
where the $\rvz_0$ is proposed to be sampled from variational posterior $q_\enc(\rvz|\rvx)$, thus the $\rvz_{K}$ can be considered as calibrated $\rvz$. For simplicity, we denote $\rvz^K = \operatorname{MCMC}_{\dec,\prior}^K(\Tilde{\rvz}|\rvz)$. Note the $\nabla_\rvz \log p_{\dec,\prior}(\rvz|\rvx)$ can be easily obtained by following form: $\nabla_\rvz \log p_{\dec,\prior}(\rvz|\rvx) = \nabla_\rvz \log p_{\dec,\prior}(\rvx,\rvz) - \nabla_\rvz\log p_{\dec,\prior}(\rvx) = \nabla_\rvz \log p_{\dec,\prior}(\rvx,\rvz)$.

Once the MCMC Calibrated $\rvz_K$ is obtained, we can conduct the constrained learning formulation similar to \cref{sec:energy calibration}. We directly give the corresponding saddle-point problem: 
\begin{equation*}
\small 
\begin{split}
    &\gL(\phi = \{\enc,\dec,\prior\}) = - \E_{p_d(\rvx)}[\mathrm{ELBO}_\phi(\rvx)]+ \lambda_1 \cdot \gL_{\mathrm{con}}(\beta) 
    + \lambda_2 \cdot \gL_{\mathrm{con}}(\enc), \quad \lambda_1, \lambda_2 \geq 0,
\end{split}
\end{equation*}
where $\gL_{\mathrm{con}}(\enc) 
    =\E_{q_{\enc}(\rvz|\rvx)}\big[\E_{p_{\dec,\prior}(\Tilde{\rvz}|\rvz,\rvx)}[ ||\rvz-\Tilde{\rvz}||_2^2]\big]$.
Given the saddle-point formulation, the learning and calibration of the encoder parameterized by $\enc$ can be achieved with corresponding empirical optimization as follows:
\begin{equation*}
\small
\begin{split}
\enc_{t+1} := \argmin_\enc \bigg\{&\sum\limits_{i=1}^{n}\sum\limits_{j=1}^{m} \Big( \log\tfrac{q_{\enc_{t}}(\rvz_{ij}|\rvx_i)}{p_{\prior_{t}}(\rvz_{ij})} -\log p_{\dec_{t}}(\rvx_i|\rvz_{ij}) \\
& + \lambda_2 \times ||\rvz_{ij}- \mathrm{MCMC}_{\dec,\prior}^{K}(\Tilde{\rvz}_{ij}|\rvz_{ij})||_2^2 \Big) \bigg\},
\end{split}
\end{equation*}
The update of the rest parameters $\{ \dec,\prior,\energy \}$ remain the same, while the $\lambda_i$ (i=1,2) is updated by \cref{eq:dual step}. 

\subsection{Energy-Calibrated Normalizing Flow}
The Energy-Calibrated Normalizing Flow (EC-Flow) can be easily formed, as normalizing flow $h_\phi$ directly models the data likelihood $p_\phi(\rvx) = p_{z}(\rvz)\times \left\vert\det\left(\frac{\partial h_\phi}{\partial \rvz}\right)\right\vert$, where $ h_\phi$ is the invertible transformation, thus we just need to use the negative log-likelihood $-\E_{p_d(\rvx)}[\log p_\phi(\rvx)]$ estimated by normalizing flow to serve as the $\gL(\phi)$, and directly sampling from $p_\phi(\rvx)$ to obtain samples, while other components keep the same.

\subsection{Applying to Zero-Shot Image Restoration}
\label{sec:image restoration method}
In this section, we present the application of our method in zero-shot image restoration tasks, inspired by recent works~\cite{wang2023zeroshot} that employ the diffusion model and the Range-Null space theory for similar purposes.

We start by briefly reviewing the necessary background. Given a known linear operator $\mA \in \R^{D_1\times D_2}$,  there exists pseudo-inverse $\mA^\dagger \in \R^{D_2\times D_1}$ that satisfies $\mA\mA^\dagger\mA = \mA$.
Considering degraded image: $\rvy = \mA\rvx$. For any prediction $\hat{\rvx}_r$, we let $\hat{\rvx} = \mA^\dagger\rvy + (\mI-\mA^\dagger\mA)\hat{\rvx}_r$, then immediately gives: $\mA\hat{\rvx} = \mA\mA^\dagger\rvy + \mathbf{0} = \mA\rvx$, which we make predicted images have the same degradation as original images. This can be regarded as $\rvx_r$ predicting the zero-space while remaining the original rang-space $\mA\rvx$. For getting the $\hat{\rvx}_r$, we can employ $\mA^\dagger\rvy$ which is the range-space part of $\rvx$ as biased ground truth and build a joint distribution 
\begin{equation}
\small
\label{eq:joint density}
    p_{\dec,\prior}(\mA^\dagger\rvy, \rvz) = p_\prior(\rvz)p(\mA^\dagger\rvy|\mA^\dagger\mA\decoder(\rvz)).
\end{equation}
However the $p_{\prior}(\rvz)$ may not be powerful enough to handle image restoration. Recall that we have an EBM in data space, hence we can construct a more powerful prior via neural transport, i.e.,
\vspace{-1mm}
\begin{equation}
\small
\label{eq:neural transport prior}
\begin{split}
    p_{\prior,\energy}(\rvz)&\propto \mathrm{exp}(-\energyfn(\decoder(\rvz)))p_{\prior}(\rvz) 
    \propto \mathrm{exp}(-E_{\dec,\energy}(\rvz))p_{\prior}(\rvz).
\end{split}
\end{equation}
This is achieved by transporting $\rvz$ to data space via decoder, then the EBM defined in data space is used, finally, the EBM defined in latent space for enhancing prior is well-constructed.
By substituting the prior in \cref{eq:neural transport prior} into \cref{eq:joint density}, the new joint likelihood is obtained. Then we run MCMC on $\rvz$ to maximize the biased joint likelihood, i.e.,
\begin{equation}
    \small
    g_\beta(\rvz_r^{k+1}) = \decoder(\rvz_r^{k}) + \frac{s}{2}\nabla_{\rvz_r^k} \log p_{\prior,\dec,\energy}(\mA^\dagger\rvy,\rvz_r^{k}) + \sqrt{s}\xi,
\end{equation}
where $\xi \sim \mathcal{N}(0,\mI)$, $\rvz_r^0$ can be initialized by sampling from $p_{\prior}(\rvz)$. In practice, the linear degraded operator $\mA$ has various corresponding image restoration tasks, such as colorization, super-resolution, and inpainting. See Appendix~\ref{app:detailed linear degraded operator} for concrete forms of $\mA$ and $\mA^\dagger$. It's worth noting that our proposed method doesn't need extra training for those tasks.

\section{Experiment}
In this section, we conduct comprehensive experiments to evaluate the proposed \shortname. We also evaluate the extension of Energy-based calibration to normalizing flow and variational inference.
We use a simple ResNet~\cite{resnet} similar to used in VAEBM~\cite{vaebm} or CLEL~\cite{clel} as energy functions $\energyfn$ with 30 MCMC steps on CIFAR-10 and 15 MCMC steps on other datasets in all experiments. We adopt Fréchet Inception Distance (FID)~\cite{heusel2017gans} as quantitative metrics in most experiments. We apply Exponential Moving Average (EMA) with coefficient of 0.999 on Church-64 and 0.9999 on other datasets for VAE. Please note that by default, our proposed model does not incorporate the use of MCMC in test time, unless specifically indicated otherwise. 
The prior $p_\prior(\rvz)$ is a simple Gaussian as default.

\captionsetup[subfigure]{labelformat=parens}

\begin{figure*}[t]
\centering
\hfill
\begin{subfigure}{.22\columnwidth}
  \centering
  \includegraphics[width=1\linewidth]{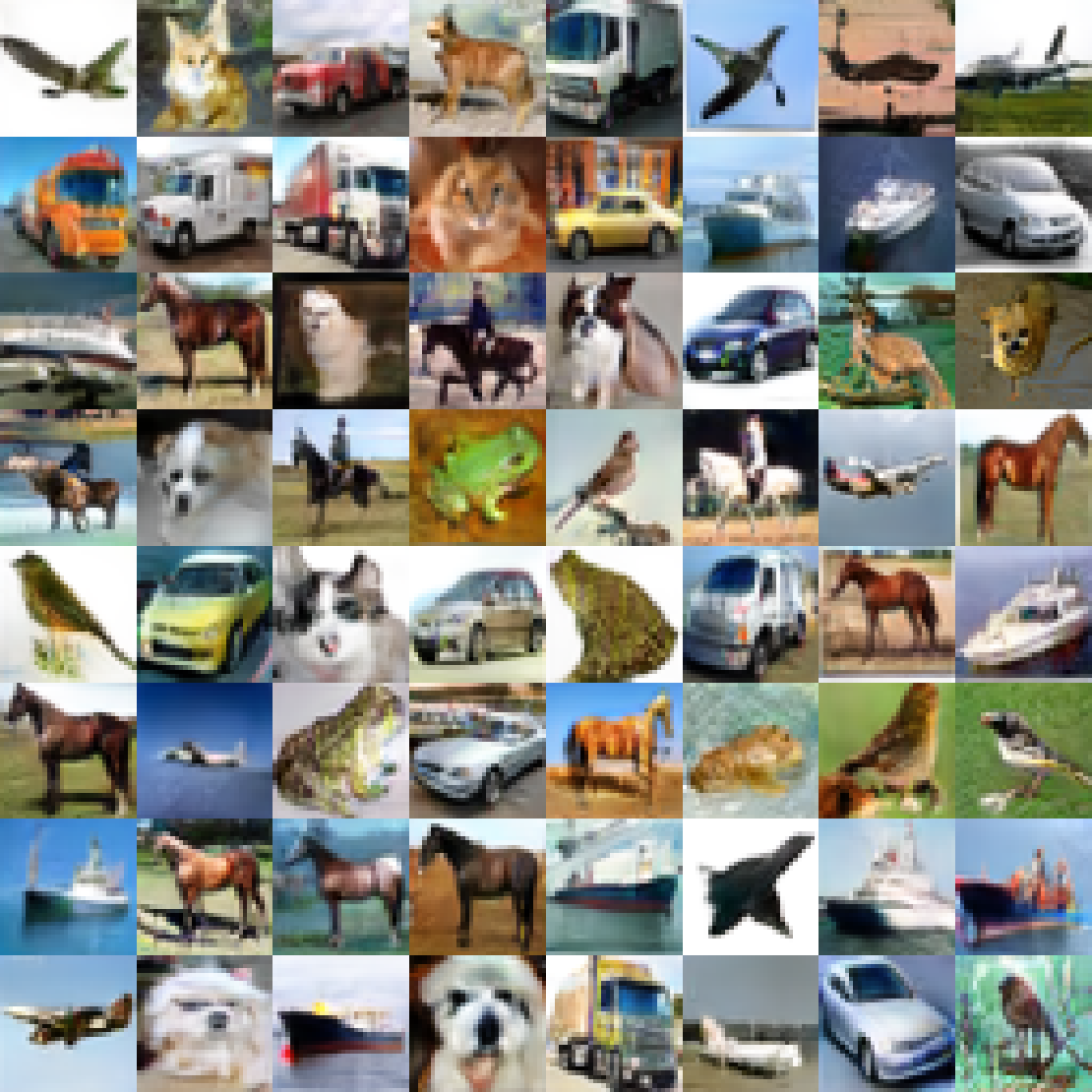}
  \caption{CIFAR-10}
\end{subfigure}
\hfill
\begin{subfigure}{.22\columnwidth}
  \centering
  \includegraphics[width=1\linewidth]{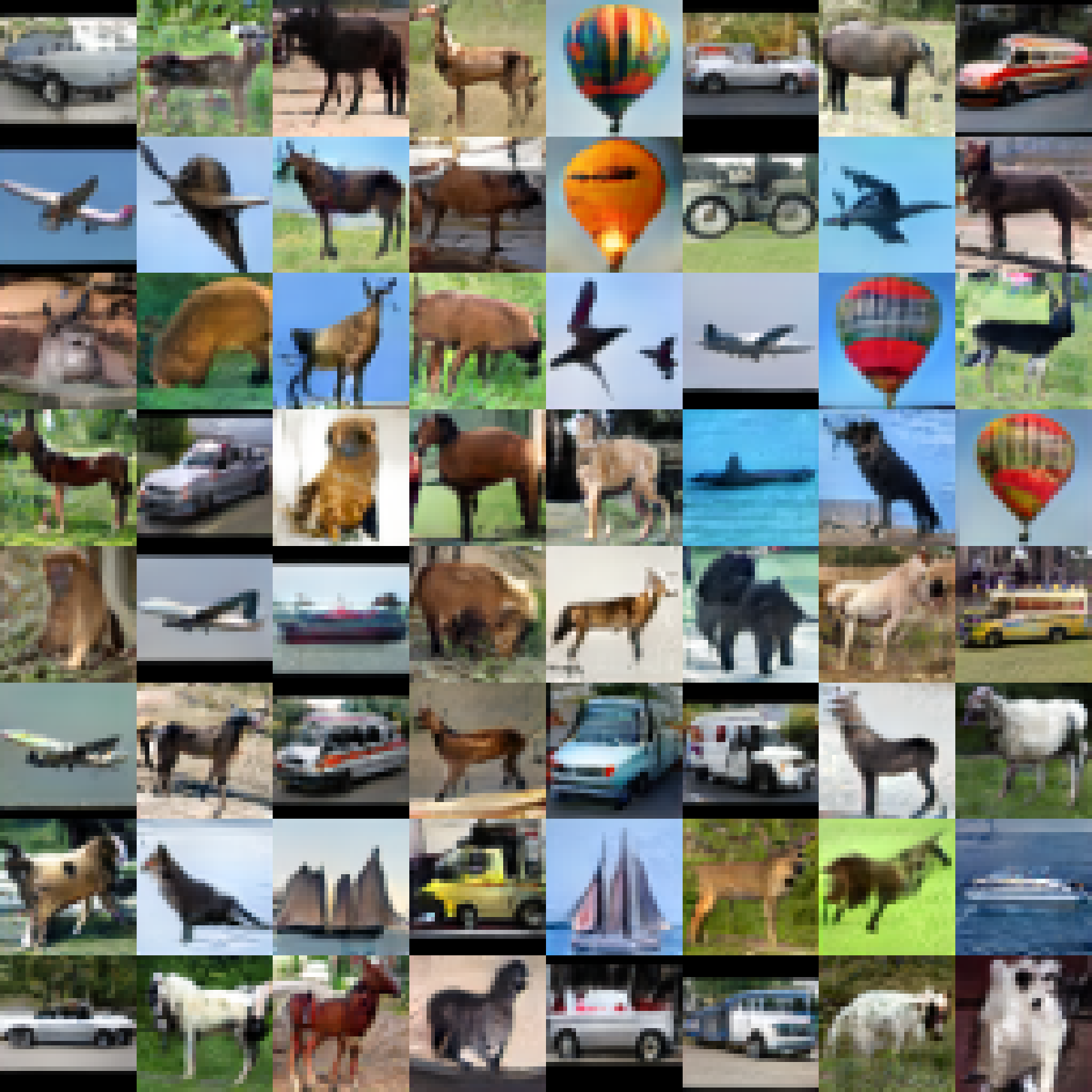}
  \caption{STL-10}
\end{subfigure}
\hfill
\begin{subfigure}{.22\columnwidth}
  \centering
  \includegraphics[width=1\linewidth]{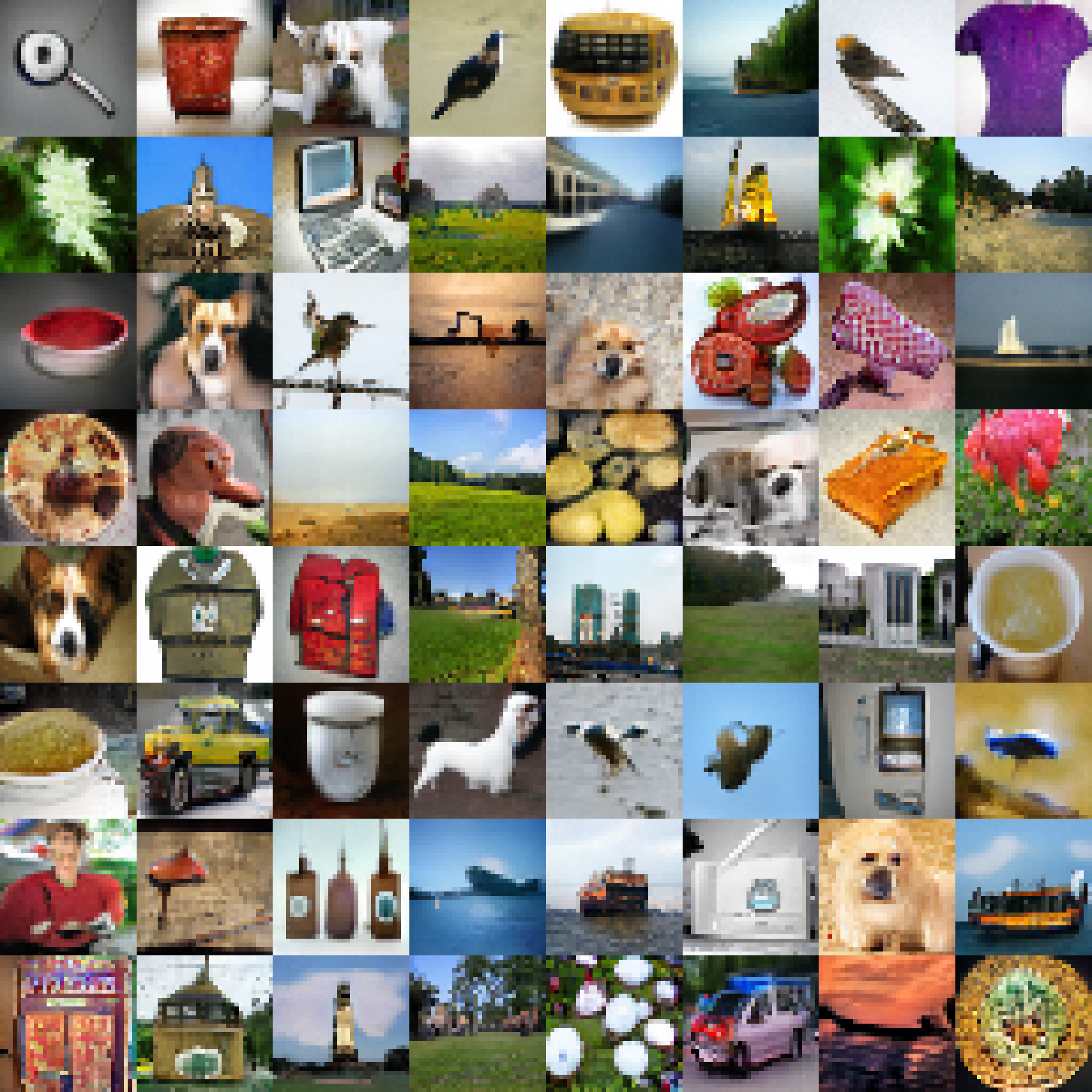}
  \caption{ImageNet 32}
\end{subfigure}
\hfill
\begin{subfigure}{.22\columnwidth}
  \centering
  \includegraphics[width=1\linewidth]{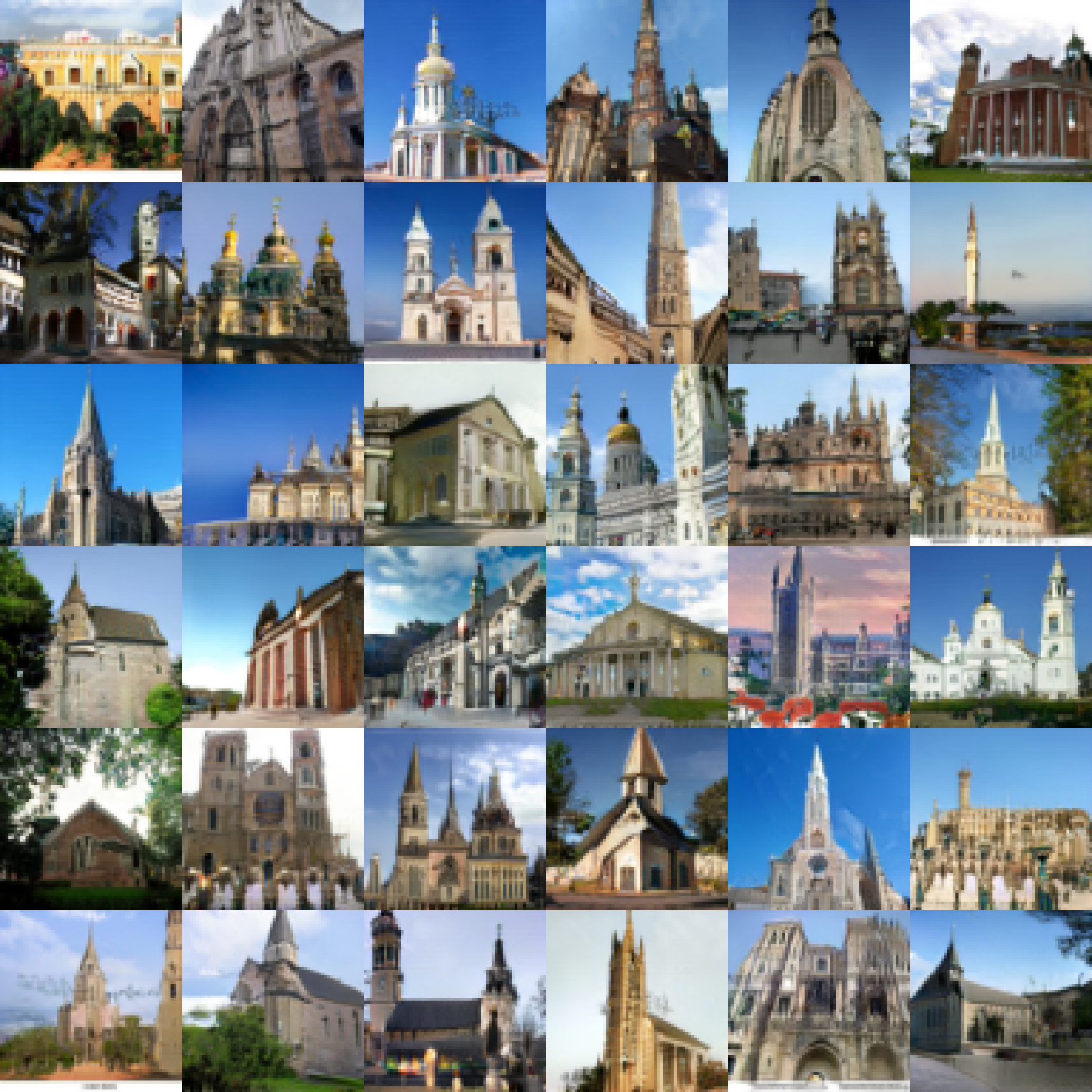}
  \caption{LSUN Church 64}
\end{subfigure}%
\vfill
\begin{subfigure}{.225\columnwidth}
  \centering
  \includegraphics[width=\linewidth]{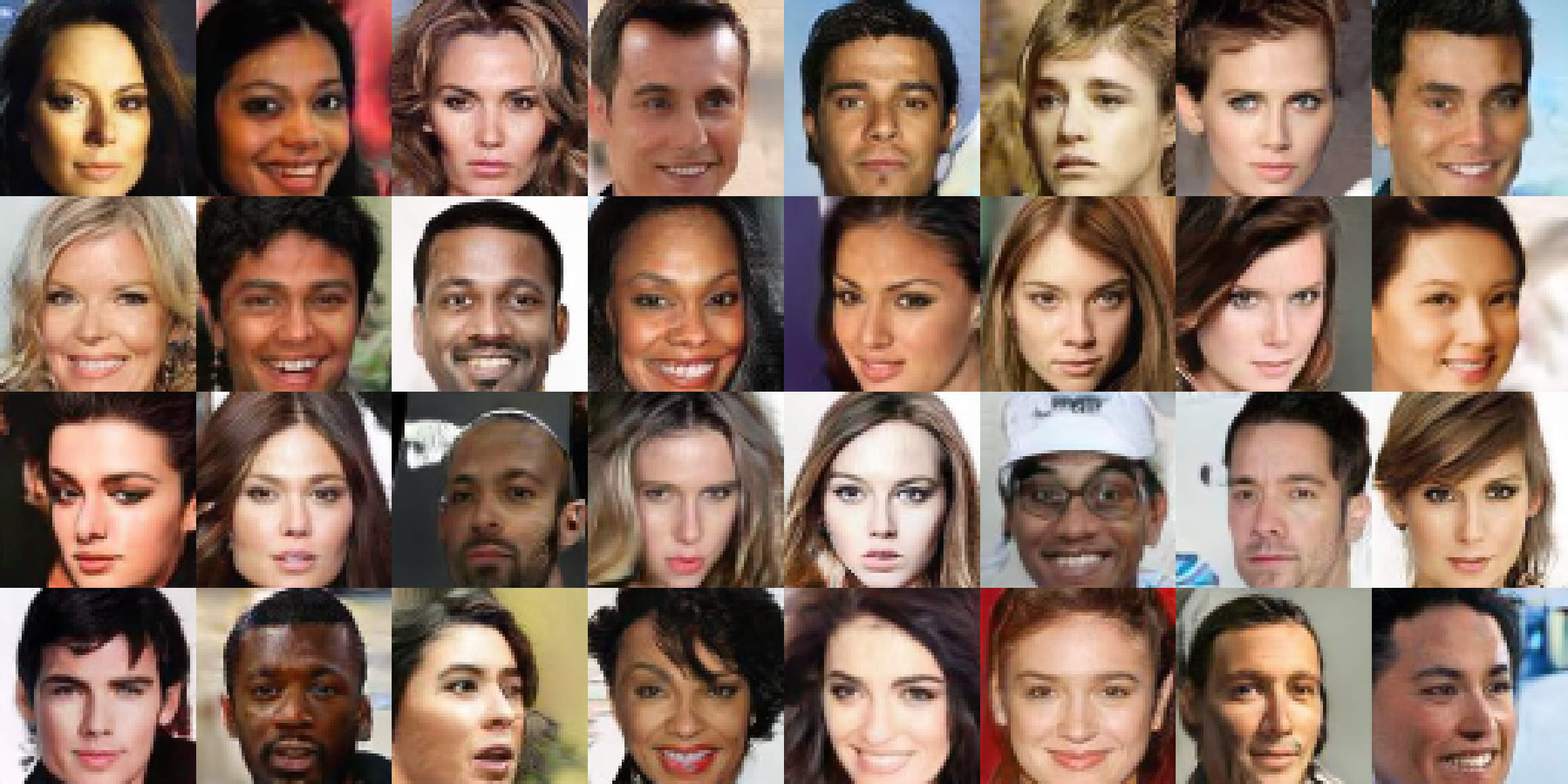}
  \caption{CelebA 64}
\end{subfigure}
\begin{subfigure}{0.67\columnwidth}
  \centering
  \includegraphics[width=\linewidth]{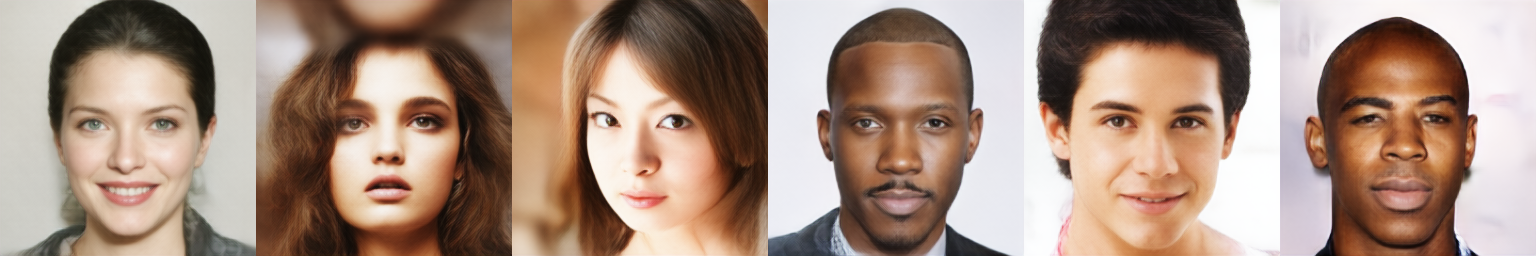}
  \caption{CelebA-HQ-256}
  \label{fig:sub5}
\end{subfigure}
\vspace{-3mm}
\caption{Random generated samples from \shortname. For CelebA 64 and CelebA-HQ-256, we pick out samples for diversity.}
\label{fig:generative images}
\vspace{-6mm}
\end{figure*}

\subsection{Image Generation}
In this section, we evaluate \shortname on six datasets, including CIFAR-10~\cite{yu2015lsun}, STL-10~\cite{STL10}, ImageNet 32~\cite{imagenet,imagenet_32}, LSUN Church 64~\cite{yu2015lsun}, CelebA 64~\cite{liu2015deep}, and CelebA-HQ-256~\cite{liu2015deep}. The backbone similar to that proposed in FastGAN~\cite{liu2020towards} is used in experiments. We use a flow as the prior in experiments on the CelebA-HQ-256.
See \cref{app:image_geneartion} for more experiment setting details. We show qualitative results in \cref{fig:generative images}. See \cref{app:qualitative_results} for more qualitative results and qualitative comparison to other models. 
The quantitative results are reported in \cref{tab:main_result,tab:stl,tab:imagenet,tab:Church64,tab:celeba64,tab:celeba256}, with the best results for GANs or Score-models highlighted by underlining, and the best results for other models in bold, respectively.

Our results are comparable to advanced GANs and Score-based Models, outperforming NVAE-family (including NVAE, NCP-VAE and VAEBM) and Glow, the most advanced Hierarchical VAE and flows, respectively, by a significant margin on all datasets, despite using much smaller latent space, and much fewer training resources. Our results also outperform existing EBMs, with a \textit{single-step generation manner}. Additionally, Our results outperform the Consistency Models with Learned Perceptual Image Patch Similarity (LPIPS)~\cite{zhang2018perceptual} which is a strong model on single-step non-adversarial generation. Note our approach does not rely on LPIPS, which is known to potentially manipulate the FID metric~\cite{imagenet_fid}.
Remarkably, we obtain a strong result of unconditional generation on ImageNet 32, even outperforming DDPM~\cite{ddpm}. This indicates the strong potential and capability of our proposed model for learning highly diverse datasets.

\begin{table}[!t]
\caption{Generative performance on CIFAR-10. The $\dagger$ means without extra hyper-parameters tuning. The * means we evaluate the FID by officially released pre-trained checkpoint.} 
\vspace{-2mm}
\label{tab:main_result}
\centering
\begin{small}
\resizebox{0.7\linewidth}{!}
{%
\begin{tabular}{llccc}
\toprule
& \textbf{Model} & NFE$\downarrow$ & \textbf{FID}$\downarrow$ & \textbf{Times ($s$)}$\downarrow$\\
\midrule
\multirow{4}{*}{\textbf{Score-based}}
& NCSN \cite{song2019generative} & 1000 & 25.32 & 107.9 \\
~ & Denoising Diffusion \cite{ho2020denoising} & 1000 & \underline{3.17} & 80.5\\ 
~ & Consistency Models (LPIPS) \cite{song2023consistency} & 1 & 8.70 & - \\ 
\midrule
\multirow{3}{*}{\textbf{GAN-based}} & SN-GAN \cite{miyato2018spectral} & 1  & 21.7 & -  \\
~ & AutoGAN \cite{gong2019autogan} & 1   & 12.4 & - \\
~ &  StyleGAN2 w/o ADA \cite{karras2020training} & 1  & 9.9  & 0.04 \\
\midrule
\multirow{2}{*}{\textbf{VAEs+GANs}} & VAE+GAN~\cite{dcvae}& 1  & 39.8 & - \\
~ & DC-VAE~\cite{dcvae} & 1   &  17.9 & - \\
\midrule
\multirow{2}{*}{\textbf{EBMs+GANs}} & FlowCE~\cite{gao2020flow}& 1   & 37.3 & -\\
~ & Divergence Triangle~\cite{han2020joint} & 1 & 30.1 & - \\
\midrule
\midrule
\multirow{2}{*}{\textbf{Flow-based }}
&Glow~\cite{kingma2018glow} & 1  & 48.9 & - \\
~&SurVAE Flow~\cite{survae}& 1  & 49.03 & - \\
\midrule
\multirow{2}{*}{\textbf{NVAE-family }} 
&NVAE \cite{vahdat2020NVAE}  & 1 & 50.97 & 0.36 \\
~&NCP-VAE \cite{nvp-vae} & -  & 24.08 & - \\
\midrule
\multirow{5}{*}{\textbf{Energy-based}} 
&IGEBM~\cite{du2019implicit} & 60  & 40.58 & -\\
~ & CoopFlow~\cite{coopflow} & 31 & 15.80~(26.54$^{\dagger *}$) & - \\
~ & CoopFlow w/o MCMC & 1 & 79.45$^*$ & -\\
~ & Hat EBM~\cite{hatebm} & 51 & 19.30 & - \\
~ & NT-EBM w/ Flow~\cite{ntebm} & - &  78.12 & - \\
~ & CLEL-Large~\cite{clel} & 1200 & 8.61 & - \\
~ & Dual MCMC~\cite{dualmcmc} & 31 & 9.26 & - \\
~ & Diffusion EBM~\cite{gao2020flow} & 180 & 9.58 & - \\
~ & VAEBM~\cite{vaebm} & 16  & 12.19 & 8.79\\
\midrule
\multirow{2}{*}{\textbf{Ours}}
& \textbf{\shortname}  & 1 & \textbf{5.20} & \textbf{0.03} \\
~&\textbf{EC-Flow}   & 1 & 31.12 & 0.37 \\
\midrule
\multirow{4}{*}{\textit{\textbf{Ablations}}} & 
\textbf{\shortname w/ MCMC} & 31  & 5.63 & 0.52\\
~&\textbf{EC-Flow w/ MCMC}  & 31 & 22.74 & 0.86 \\
~ & \textbf{VAE}   & 1 & 104.68 & 0.03 \\
~&\textbf{Flow}   & 1 & 91.33 & 0.37\\
\bottomrule
\end{tabular}
}
\end{small}
\vspace{-4mm}
\end{table}

\begin{figure}[!t]
    \centering
    \vspace{1.5mm}
    \includegraphics[width=0.5\linewidth]{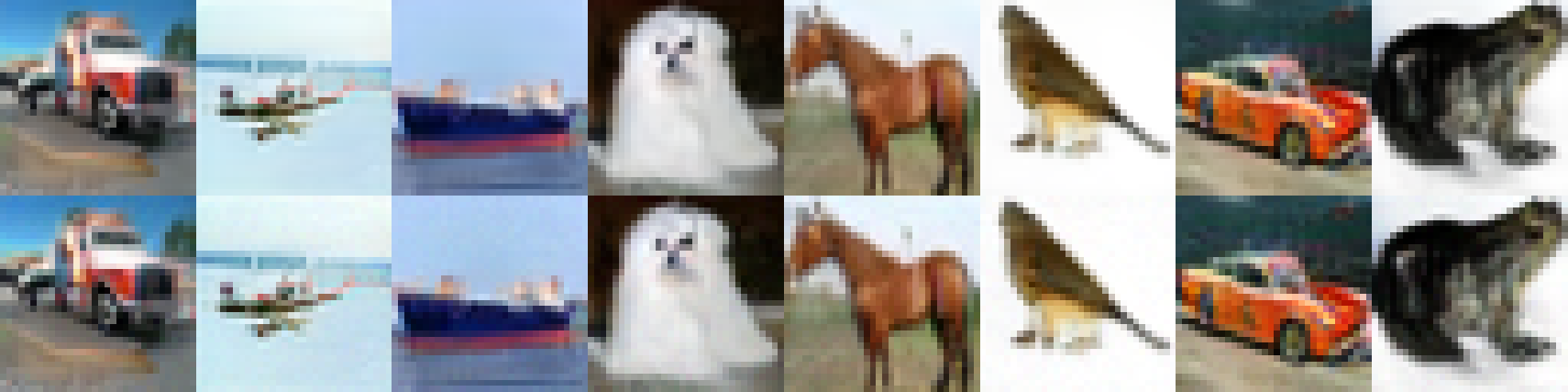}
    \vspace{-3mm}
    \caption{Comparison of \textbf{\shortname}~(Top) and \textbf{\shortname (w/ MCMC)}~(Bottom) on CIFAR-10. 
    Best viewed when zoomed in.
    }
    \label{fig:comparison_noise}
    \vspace{-4mm}
\end{figure}

\begin{table}[t]
\centering
\begin{adjustbox}{valign=m} 
\begin{minipage}[t]{0.32\linewidth}
    \centering
    \caption{\small Generative performance on STL-10.}\label{tab:stl} 
    \vspace{-3mm}
    \scalebox{0.8}{
        \begin{tabular}{@{}lc@{}}
            \toprule
            \textbf{Model} & \textbf{FID$\downarrow$} \\
            \midrule
            ProbGAN~\cite{he2019probgan}   & 46.7 \\
            SN-GAN~\cite{miyato2018spectral}    & 40.1  \\
            Improv. MMD GAN~\cite{wang2019improving} & 37.6 \\
            AutoGAN~\cite{gong2019autogan}   & \underline{31.0} \\
            DC-VAE~\cite{dcvae}    & 41.91\\
            \midrule
            \midrule
            \textbf{\shortname (Ours)} & \textbf{8.39} \\
            \bottomrule
        \end{tabular}
    }
\end{minipage}
\end{adjustbox}
\hfill
\begin{adjustbox}{valign=m} 
\begin{minipage}[t]{0.32\linewidth}
    \centering
    \caption{\small Generative performance on ImageNet 32.} \label{tab:imagenet}
    \vspace{-3mm}
    \scalebox{0.8}{
        \begin{tabular}{@{}lc@{}}
            \toprule
            \textbf{Model} & \textbf{FID$\downarrow$} \\
            \midrule
            DDPM~\cite{ddpm} & 6.99 \\
            Flow Matching~\cite{lipman2023flow} & \underline{5.02} \\
            \midrule
            \midrule
            PixelCNN~\cite{oord2016conditional} &  40.51\\
            \midrule
            CF-EBM~\cite{zhao2021learning} & 26.31 \\ 
            CLEL-Large~\cite{clel} & 15.47 \\
            \midrule
            \textbf{\shortname (Ours)} & \textbf{5.76} \\
            \bottomrule
        \end{tabular}
    }
\end{minipage}
\end{adjustbox}
\hfill
\begin{adjustbox}{valign=m} 
\begin{minipage}[t]{0.32\linewidth}
    \centering
    \caption{Generative performance on Church 64.}\label{tab:Church64} 
    \vspace{-3mm}
    \scalebox{0.8}{
        \begin{tabular}{@{}lc@{}}
            \toprule
            \textbf{Model} & \textbf{FID$\downarrow$ } \\
            \midrule
            NVAE~\cite{vahdat2020NVAE} & 41.3 \\
            GLOW~\cite{kingma2018glow} & 59.35 \\
            \midrule
            Diffusion EBM~\cite{gao2020learning} & 7.02 \\
            Dual MCMC~\cite{dualmcmc} & 4.56 \\
            VAEBM~\cite{vaebm} & 13.51 \\
            \midrule
            \textbf{\shortname (ours)} & \textbf{4.28} \\
            \bottomrule
        \end{tabular}
    }
\end{minipage}
\end{adjustbox}
\\
\vspace{1mm}
\begin{adjustbox}{valign=m} 
\begin{minipage}[t]{0.34\linewidth}
    \centering
    \caption{\small Generative performance on CelebA 64.}\label{tab:celeba64} 
    \vspace{-2.mm}
    \scalebox{0.8}{
        \begin{tabular}{@{}lc@{}}
            \toprule
            \textbf{Model} & \textbf{FID$\downarrow$}\\
            \midrule
            NCSNv2~\cite{song2020improved} & 26.86 \\
            \midrule
            COCO-GAN~\cite{lin2019coco} & \underline{4.0} \\
            QA-GAN~\cite{parimala2019quality} & 6.42 \\
            Divergence Triangle~\cite{han2020joint} & 24.7 \\
            \midrule
            \midrule
            Dual MCMC~\cite{dualmcmc} & 5.15 \\
            VAEBM~\cite{vaebm} & 5.31\\
            NCP-VAE~\cite{nvp-vae} & 5.25 \\
            NVAE~\cite{vahdat2020NVAE} & 14.74\\
            \midrule
            \textbf{\shortname (Ours)} & \textbf{2.71} \\
            \bottomrule
        \end{tabular}
    }
\end{minipage}
\end{adjustbox}
\hspace{3mm}
\begin{adjustbox}{valign=m} 
\begin{minipage}[t]{0.34\linewidth}
    \centering
    \caption{\small Generative performance on CelebA-HQ-256.} \label{tab:celeba256}
    \vspace{-2.mm}
    \scalebox{0.8}{
        \begin{tabular}{@{}lc@{}}
            \toprule
            \textbf{Model} & \textbf{FID$\downarrow$} \\
            \midrule
            ALAE~\cite{pidhorskyi2020adversarial} & 19.21\\
            DC-VAE~\cite{dcvae} & 15.81\\
            PGGAN~\cite{karras2017progressive} & \underline{8.03}\\
            \midrule
            \midrule
            GLOW~\cite{kingma2018glow}&68.93\\
            \midrule
            Dual MCMC~\cite{dualmcmc} & 15.89 \\
            VAEBM~\cite{vaebm}& 20.38\\
            NCP-VAE~\cite{nvp-vae} & 24.79 \\
            NVAE~\cite{vahdat2020NVAE} & 45.11\\
            \midrule
            \textbf{\shortname (Ours)} & \textbf{12.35} \\
            \bottomrule
        \end{tabular}
    }
\end{minipage}
\end{adjustbox}
\vspace{-3mm}
\end{table}

\spara{Comparison with Other EBMs} The proposed method presents a crucial, clear difference from existing methods: we demonstrate that it is possible to drop MCMC steps during test time sampling without compromising the quality of generation. And the FID is even better without MCMC. This is likely because MCMC introduces noise at each step which cannot be perfectly denoised~(See \cref{fig:comparison_noise}), resulting in noise still present in the image.  In contrast, the decoder minimizes the distance $\E_\rvx ||\Tilde{\rvx},\rvx||_2^2$, assuming $\Tilde{\rvx}= \mathbf{y} + \xi$, where $\xi$ is zero-mean random noise without any information. The minimum is ideally achieved at $\rvx = \mathbf{y}$, allowing the decoder to act as a filter and output sharp images without noise. 
Moreover, as observed, most previous EBMs~\cite{du2019implicit, vaebm,coopflow} need to tune MCMC steps and step size carefully during inference to achieve high performance~(e.g., CoopFlow$^\dagger$ in \cref{tab:main_result}), while our method does not need extra hyper-parameters tuning as even no MCMC required during inference.

\spara{Energy-Calibrated Normalizing Flow}
We evaluate the EC-Flow on CIFAR-10. We use Glow as the flow architecture. As shown in \cref{tab:main_result}, the EC-Flow outperforms existing flows in single-step generation, even outperforming the FlowCE which is trained by playing an adversarial game. 
However, we found that in the case of EC-Flow, the role of MCMC remains crucial in enhancing generative performance. 
Specifically, EC-Flow w/ MCMC significantly improves the FID from 31.12 to 22.74. 
We believe that this is due to the expressivity limitations of invertible transformations in flows.
Nevertheless, compared to CoopFlow which also combines flows and EBMs, our EC-Flow beat CoopFlow without MCMC by a large margin, demonstrating the effectiveness of the proposed EC-Flow. Note that the 15.8 FID achieved by CoopFlow needs extra hyper-parameter tuning, its result (26.54 FID) without extra tuning is worse than ours w/ MCMC (22.74 FID).

\spara{Energy-Calibrated Variational Learning}
We also evaluate Energy-Calibrated Variational Learning on CIFAR-10, with 5 MCMC steps for calibrating posterior. As shown in \cref{tab:calibrated posterior}, the calibrated posterior consistently enhances the ELBO and MSE of related baseline VAE and \shortname. This indicates that the calibrated posterior can provide a more accurate posterior for better likelihood maximization. 
However, a slight FID deterioration is observed in the \shortname with Calibrated posterior, implying that the likelihood is not always consistent with generation quality.
Interestingly, we found the \shortname also improves the ELBO and MSE compared to VAE, despite we only calibrate the generative side. This may be highly due to the energy-based calibration effectively improving the ability to map latent space to data space and reducing the gap between prior and aggregated posterior.

\begin{figure}[!tbp]
\small
\centering
\includegraphics[width=.65\linewidth]{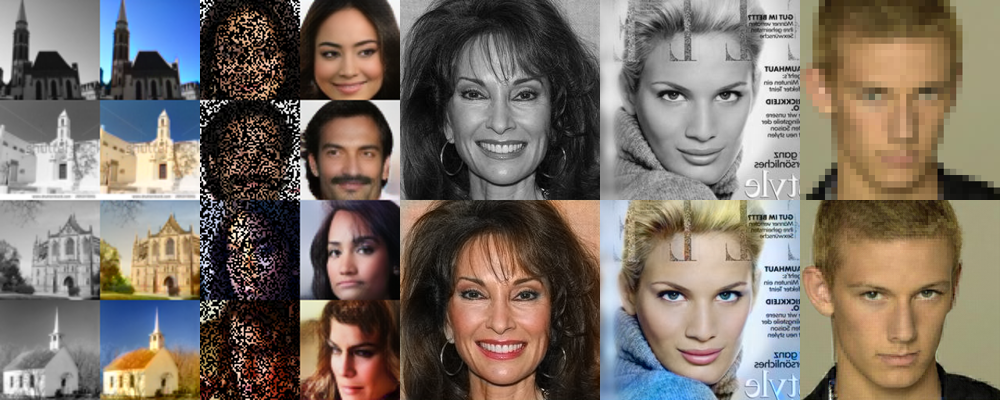}
\vspace{-2.5mm}
\caption{Qualitative results of zero-shot image restoration (colorization, inpainting, 4$\times$ super-resolution).}
\label{fig:image restoration}
\vspace{-6mm}
\end{figure}

\begin{table*}[!t]
    \begin{minipage}{.4\linewidth}
        \small
        \footnotesize
        \caption{Quantitative results Calibrated posterior on CIFAR-10.}
        \vspace{-2mm}
        \label{tab:calibrated posterior}
        \resizebox{\linewidth}{!}{
           \begin{tabular}{llll}
           \toprule
              \textbf{Model} & \textbf{FID$\downarrow$ }& \textbf{MSE$\downarrow$} & \textbf{ELBO$\uparrow$} \\
              \midrule
              VAE & 104.68 & 0.0235 & -953.43 \\
              \textbf{+ Calibrated posterior} & 102.10 & 0.0198 & -349.14 \\ 
              \midrule
              \textbf{\shortname} & \textbf{5.20} & 0.0193 & 652.59\\ 
              \textbf{+ Calibrated posterior} & 5.81 & \textbf{0.0170} & \textbf{1072.40}\\ 
        \bottomrule
        \end{tabular}
        }
    \end{minipage}
    \begin{minipage}{.6\linewidth}
        \small
            \caption{Sampling efficiency and training cost on CIFAR-10. Time is the seconds took to generate 50 images.}
            \vspace{-2mm}
        \label{tab:sampling efficiency}
        \resizebox{\linewidth}{!}
        {
            \begin{tabular}{lcccccc}
            \toprule
            Model & FID$\downarrow$ & Latent dim$\downarrow$ & Time$\downarrow$ & GPU days $\downarrow$ & GPU-Type\\
            \hline
            NCSN~ & 25.32 & 3072 & 107.9 & -  & -\\ 
            \hline
            GLOW~\cite{survae} & 48.9 & 3072 & -  & 60 & -\\
            SurVAE Flow~\cite{survae} & 49.03 & 1536  & - & 7 & TITAN-X\\
            \hline
            NVAE~\cite{vahdat2020NVAE} & 50.97 & 153600 & 0.36 & 18.3 & 32G-V100\\
            NCP-VAE~\cite{nvp-vae} & 24.08 & 153600 & -  & 34.5 & 32G-V100\\
            VAEBM~\cite{vaebm} & 12.19 & 153600 & 8.79  & $\geq$18.3 & 32G-V100\\
            \hline
           \textbf{\shortname w/ MCMC (ours)} & 5.63 & \textbf{128} & 0.52  & \textbf{3} & RTX-3090\\
           \textbf{\shortname (ours)} & \textbf{5.20} & \textbf{128} & \textbf{0.03}  & \textbf{3} & RTX-3090\\
            \bottomrule
        \end{tabular}
        }
    \end{minipage}
\vspace{-5mm}
\end{table*}

\subsection{Sampling Efficiency and Training Cost}

Although the score-based model and NVAE's variants have shown outstanding sample quality, their sampling speed is limited by the necessity of expensive MCMC sampling steps. As shown in \cref{tab:sampling efficiency}, the generation by \shortname, in contrast, takes just one pass, making it hundreds of and thousands of times faster than NCSN and VAEBM, respectively. Notably, even after performing MCMC steps, our method still only requires 0.52 seconds to generate 50 samples. This is because our energy network is lightweight and only requires 30 MCMC steps on the data space, while VAEBM runs MCMC in $(\rvx,\rvz)$ space which requires backward through its heavy decoder at each step. 
Furthermore, due to the extremely large scale of latent variables used in NVAE-family and previous flows, yielding challenges in learning, they need at least 7 GPU days to be trained on CIFAR-10, while our model only needs 3 GPU days despite using costly MCMC steps in training.

\subsection{Image Restoration}
Here we show that well-trained \shortname is able to be zero-shot used in image restoration as described in \cref{sec:image restoration method}. The Qualitative results are shown in \cref{fig:image restoration}. Our model can successfully restore those images with high quality and consistency. More qualitative results can be found in Appendix~\ref{app:qualitative_results}. 
Following the setting as in~\cite{wang2023zeroshot}, we compare our method with strong zero-shot baselines, using metrics FID, PSNR, and Consistency\cite{wang2023zeroshot}~(i.e., $||\mA \hat{\rvx}-\mA \rvx||_1$). As shown in \cref{tab:image restoration}, we outperform GAN-based PULSE~\cite{menon2020pulse} and compete with diffusion-based DDNM~\cite{wang2023zeroshot} and DDRM~\cite{ddrm}, which confirms our model provides competitive performance.

\begin{table}[!t]
\vspace{-3mm}
\centering
\begin{adjustbox}{valign=m} 
    \begin{minipage}{0.49\linewidth}
        \centering
        \footnotesize
        \caption{Quantitative results of image restoration on CelebA-HQ.}
        \vspace{-2mm}
        \scalebox{0.75}{
            \begin{tabular}{ccc}
            \toprule
            \multirow{2}{*}{\textbf{Model}} & 4$\times$ SR & Colorization\\\cmidrule(lr){2-2}\cmidrule(lr){3-3}
            &  PSNR$\uparrow$ / FID$\downarrow$ & Cons$\downarrow$ / FID$\downarrow$\\
            \midrule
            PULSE~\cite{menon2020pulse} & 22.7 / 40.3 & N/A \\
            DDRM~\cite{ddrm} & \textbf{31.6} / 31.0 & 456 / 31.2 \\ 
            DDNM~\cite{wang2023zeroshot} & \textbf{31.6} / \textbf{22.3} & 26.2 / 26.4 \\
            \midrule
            \textbf{\shortname~(ours)} & 28.8 / 30.4 & \textbf{0.004} / \textbf{13.3} \\
            \bottomrule
        \end{tabular}
        }
        \label{tab:image restoration}
    \end{minipage}
\end{adjustbox}
\hfill
\begin{adjustbox}{valign=m} 
    \begin{minipage}{0.49\linewidth}
        \centering
        \small
        \caption{Comparison for FID on CIFAR-10 between several related methods.}
        \label{tab:ablation}
        \vspace{-2mm}
        \scalebox{0.75}{
           \begin{tabular}{ll}
           \toprule
              \textbf{Model} & \textbf{FID$\downarrow$} \\
              \hline
              \textbf{\shortname} & 5.20\\ 
              \hline
              \textbf{\shortname w/ flow prior} & 4.85\\ 
              \textbf{\shortname w/ LPIPS} & 4.77\\ 
              \textbf{\shortname w/o primal-dual} & 5.72\\ 
              \hline
              VAE & 104.68 \\
              EBM & 45.52 \\
              EBM init w/ VAE samples & 44.63 \\
              VAE + WGAN & 33.78 \\ 
           \bottomrule
           \end{tabular}
        }
    \end{minipage}
\end{adjustbox}
\vspace{-3mm}
\end{table}

\subsection{Ablation Study}
\vspace{-2.0mm}
All experiments here are performed on CIFAR-10 for faster training. See Appendix~\ref{app:ablation} for Experimental settings and additional ablations. 

\spara{Effect of Calibration} The proposed method is calibrating generative models by incorporating generated samples into training. This raises the question of what is the performance of single generative model and train generative model and EBMs as independent components. We compare the proposed \shortname with three related baselines i.e., VAE, EBM, and EBM initialized with pre-trained VAE samples. As shown in Table.~\ref{tab:ablation}, we significantly outperform these three variants by a huge margin. Note that there is almost no benefit of initializing EBM with pre-trained VAE samples, implying the effectiveness of calibration.

\spara{Energy-based calibration vs.\ Adversarial-based Calibration}
The gradient for updating EBMs $\energyfn$ is similar to the gradient updates of WGAN's discriminator~\cite{wgan}. The key difference is that the EBMs draw samples from $\energyfn$ by MCMC, while WGAN draws negative samples from a generator. WGAN updates the generator and discriminator by adversarial training, while we update EBMs by maximizing the likelihood and we update the VAE by maximizing adaptive weight between the lower bound of data likelihood and conditional likelihood on calibrated samples (i.e., $\log p(\Tilde{\rvx}|\rvx) = -\frac{||\Tilde{\rvx}-\rvx||_2^2}{2\sigma^2} + \mathrm{Constant}$). Thus, a related variant of \shortname is updating the VAE by both ELBO and WGAN training objectives. We compare the adversarial variant with the proposed \shortname. As shown in Table.~\ref{tab:ablation}, the adversarial variant achieves similar results (\textbf{33.78 FID}) to the VAEs+GANs (\textbf{39.8 FID}) in DC-VAE which is significantly worse than the proposed \shortname (\textbf{5.20 FID}), highlighting the advantage of our method compared to adversarial training.

\spara{Effect of Primal-Dual} We firstly emphasize that primal-dual has advantages with theoretical guarantees in constrained learning~\cite{chamon2021constrained}. This section is for empirically showing the impact intuitively. 
As shown in \cref{tab:ablation}, in comparison to primal-dual, the naive stochastic gradient descent~(w/o dual variable $\lambda$) algorithm yields worse results. 

\spara{Compatibility with Advanced Technique in VAEs} 
We employ MSE as the distance metric and a Gaussian as prior in CIFAR-10. Recognizing the efficacy of flexible priors and better distance metrics in improving vanilla VAEs~\cite{GLF,nvp-vae}, we investigate the impact of integrating either flow prior or LPIPS into the \shortname. The superior performance of both variants~(\cref{tab:ablation}) suggests that \shortname is orthogonal to these advanced techniques to some extent. Additionally, we note that the improvement margin is not large, indicating that \shortname effectively addressed the prior hole issue and blurry generation issue in vanilla VAEs.

\begin{figure*}[!tbp]
  \centering
  \begin{subfigure}{0.24\linewidth}
    \includegraphics[width=0.97\linewidth]{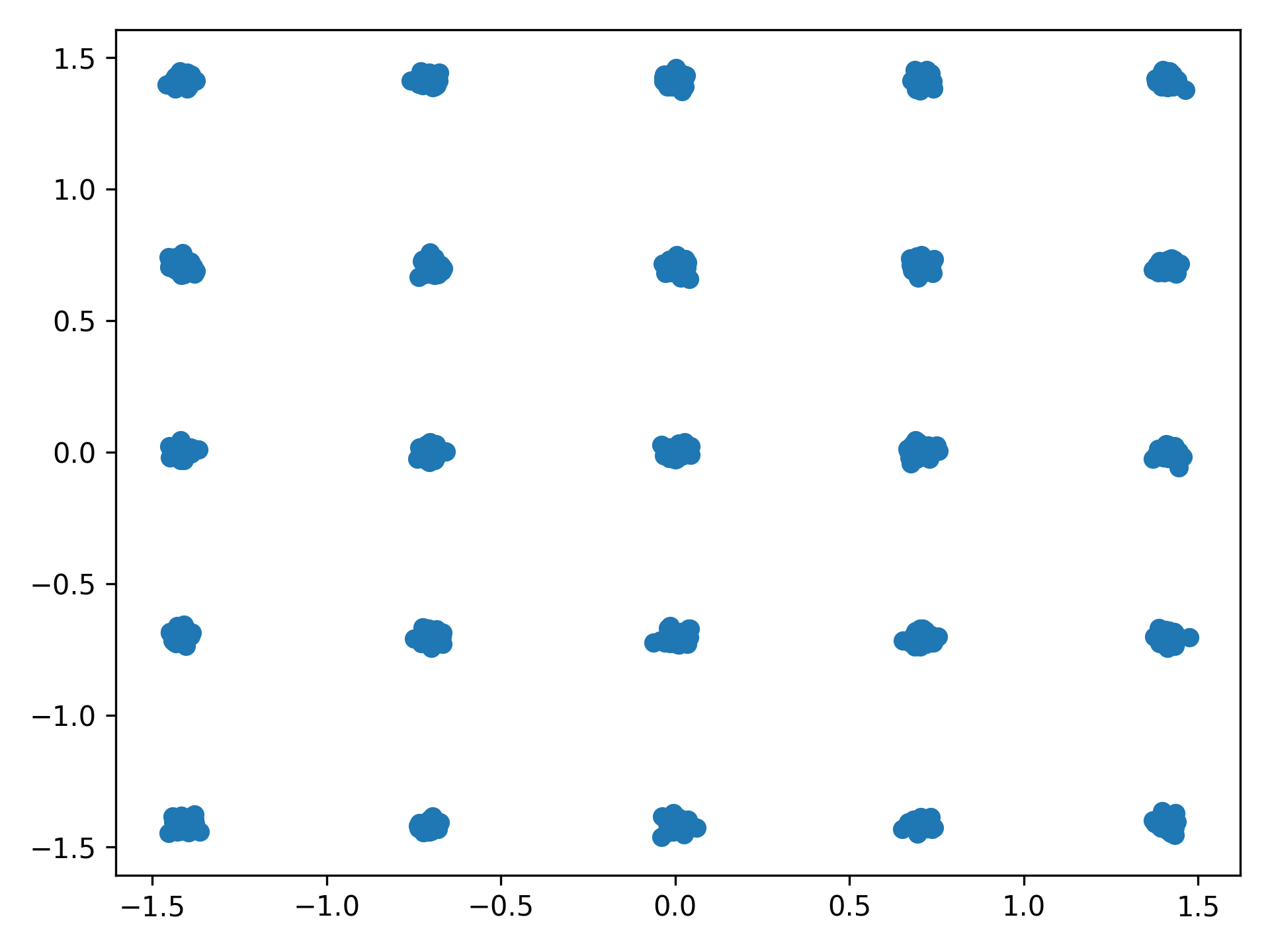}
    \captionsetup{font={tiny}}
    \caption{Real Samples}
  \end{subfigure}
  \hfill
\begin{subfigure}{0.24\linewidth}
    \includegraphics[width=0.97\linewidth]{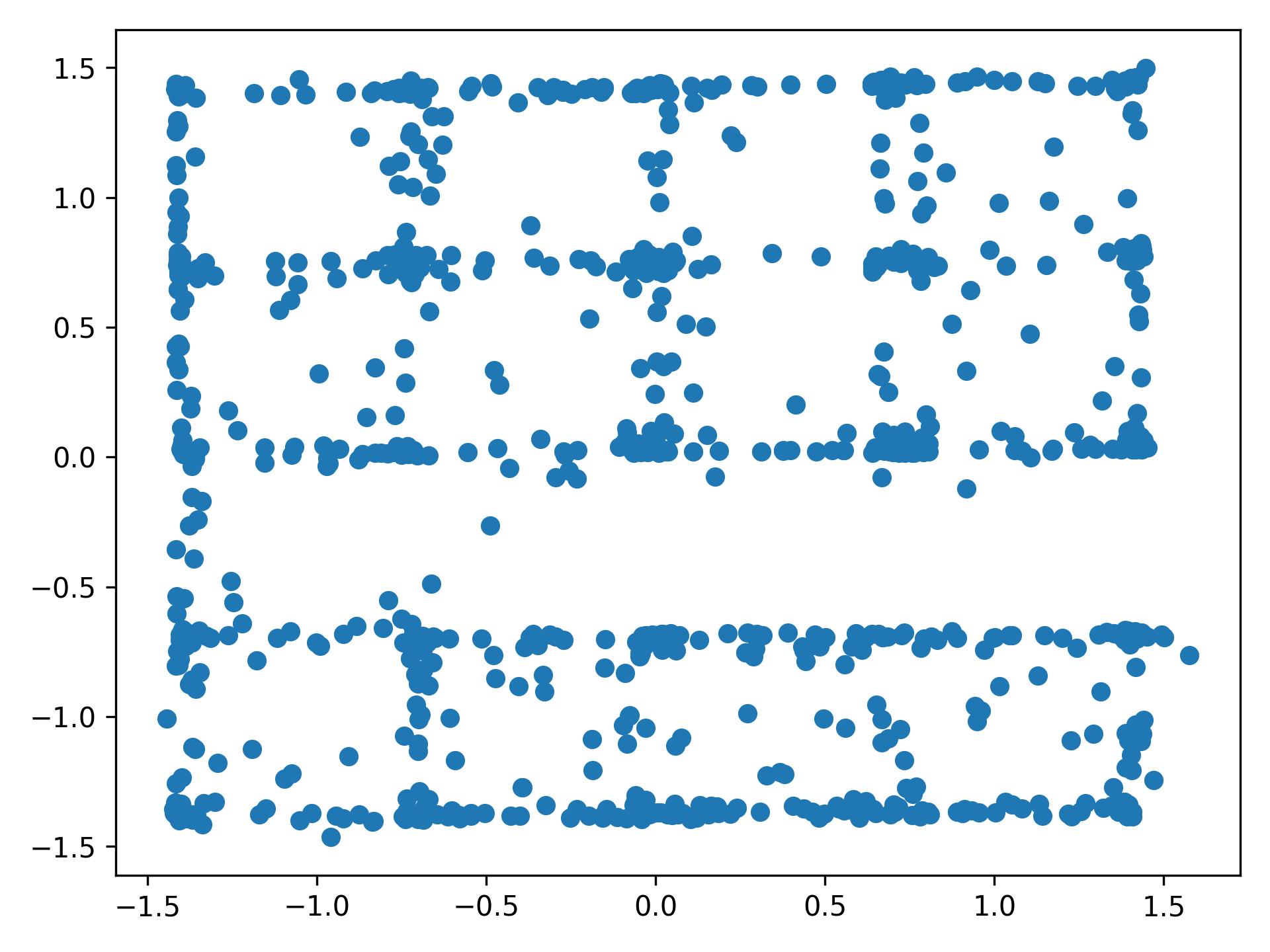}
    \captionsetup{font={tiny}}
    \caption{Samples by VAE}
  \end{subfigure}
  \hfill
\begin{subfigure}{0.24\linewidth}
    \includegraphics[width=0.97\linewidth]{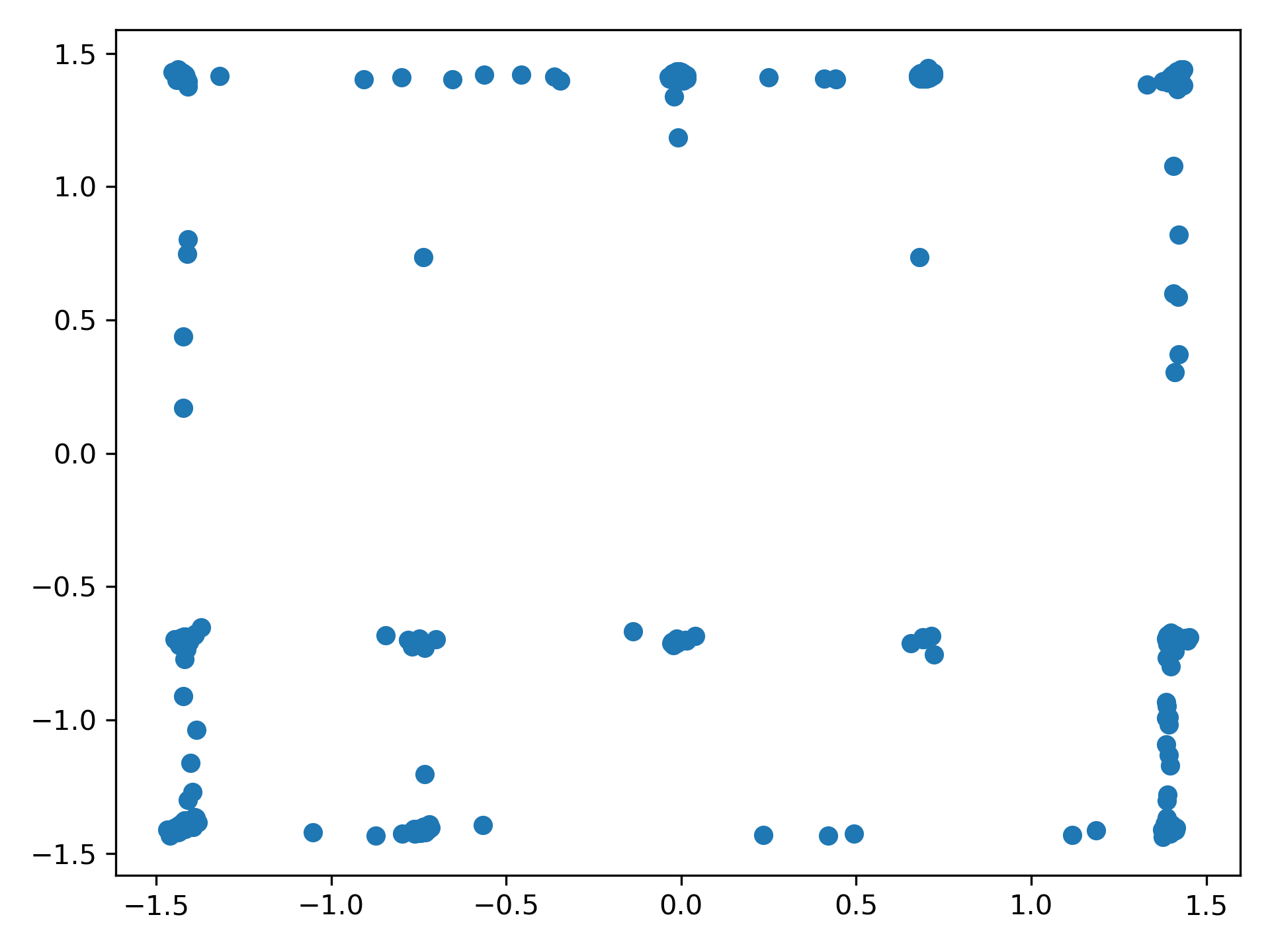}
    \captionsetup{font={tiny}}
    \caption{Samples by GAN}
  \end{subfigure}
\hfill
  \begin{subfigure}{0.24\linewidth}
    \includegraphics[width=0.97\linewidth]{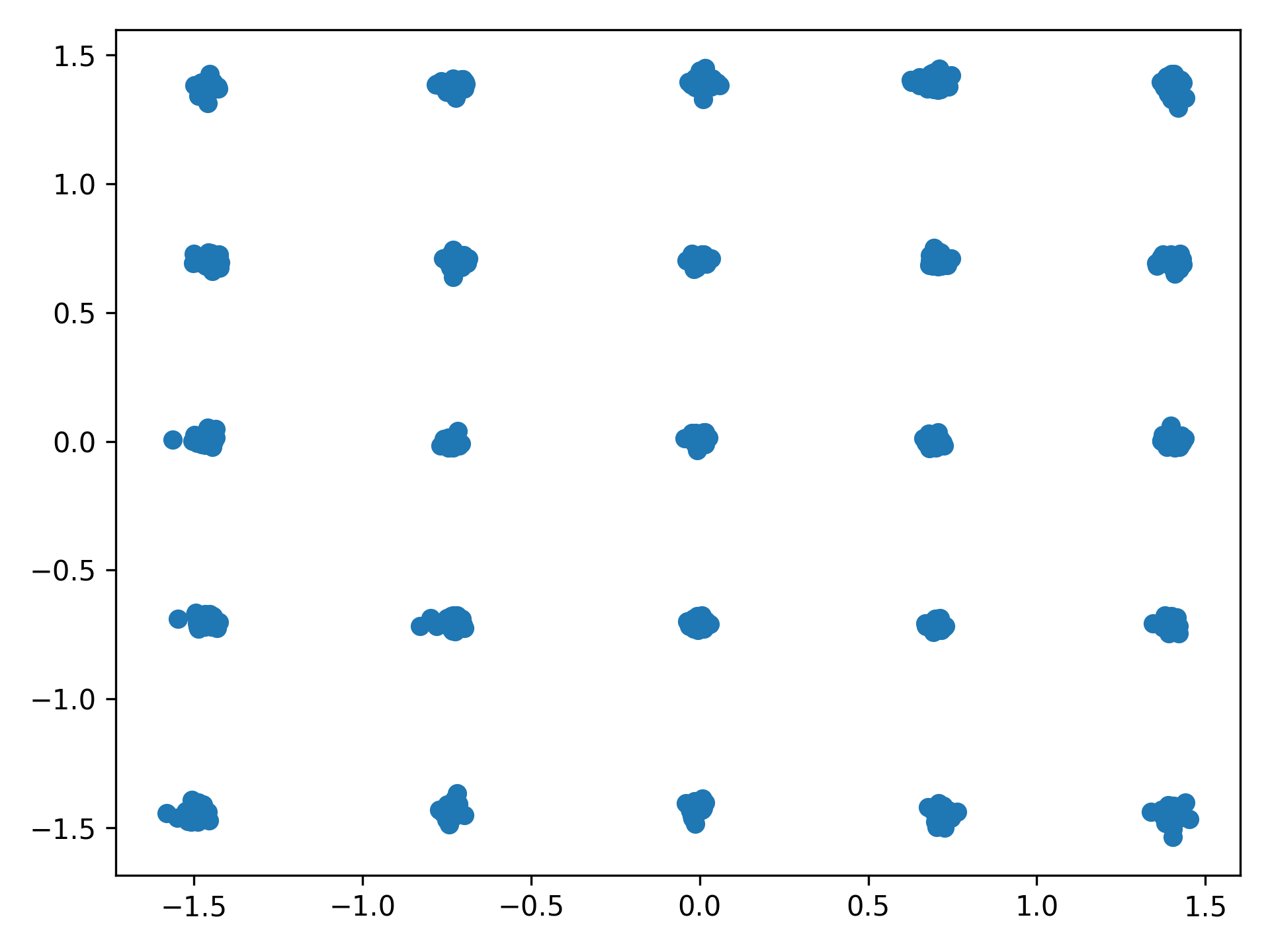}
    \captionsetup{font={tiny}}
    \caption{Samples by \shortname}
  \end{subfigure}
  \vspace{-3mm}
  \caption{Qualitative results on the 25-Gaussians dataset.}
  \label{fig:toy_data}
  \vspace{-6mm}
\end{figure*}
\subsection{Extra Study on Mode Coverage}
We evaluate the mode coverage of our model on the popular 25-Gaussians. This 2D toy dataset is also used in previous work~\cite{vaebm}. We train our \shortname with 15 MCMC steps and compare it to other models in Fig.~\ref{fig:toy_data}.
The VAE's encoder, decoder, and energy function both consist of four linear layers with 256 hidden units. The latent dimension is 20. 
We observe that the vanilla VAEs can not produce good samples, lots of samples are significantly out of modes. In contrast, our \shortname successfully calibrated the sampling distribution with all modes covered and high-quality samples. 
We also train a GAN with similar networks for comparison. We observe that GAN suffers severely from mode collapse. 
Moreover, as the true distribution is available, we also evaluate the likelihood of $100000$ generated samples by models with true data density. Our \shortname obtains \textbf{-1.07 nats} which significantly improves the likelihood obtained by VAE which is \textbf{-1.78 nats}. For reference, the likelihood of $100000$ real data under real density is \textbf{-1.04 nats}.

\vspace{-3mm}
\section{Conclusion}
\vspace{-2mm}
In this paper, we proposed \shortname, using conditional EBMs for calibrating the VAEs.
Furthermore, the proposed energy-based calibration can enhance normalizing flows and variational posterior.
We also propose and show that \shortname can effectively solve image restoration in a zero-shot manner.
We show that MCMC sampling is not required once the VAE is calibrated while keeping high performance. 
In terms of efficiency, \shortname can be trained by a single GPU and is fast to converge, addressing the intensive computational resources consumption problem of previous state-of-the-art VAEs (i.e., NVAE).  The proposed \shortname shows promising results over multiple datasets with high computational efficiency, significantly reducing the gap with GANs. 

\section*{Acknowledgments}
Jing Tang's work is partially supported by the National Natural Science Foundation of China (NSFC) under Grant No.\ U22B2060, by National Key R\&D Program of China under Grant No.\ 2023YFF0725100, by National Language Commission under Grant No.\ WT145‐39, by The Department of Science and Technology of Guangdong Province under Grant No.\ 2023A1515110131, by Guangzhou Municipal Science and Technology Bureau under Grant No.\ 2023A03J0667 and 2024A04J4454, by Hong Kong Productivity Council (HKPC), and by Createlink Technology Co., Ltd.

\bibliographystyle{splncs04}
\bibliography{main}
\clearpage
\setcounter{page}{1}
\appendix

\section{Pseudo Code of Proposed Algorithm}
\label{app:algorithm}
\begin{algorithm}[H]
    \small
    \SetKwInOut{Input}{input}
    \SetKwInOut{Output}{output}
    \DontPrintSemicolon
    \Input{Learning iterations~$T$, number of MCMC steps $K$, observed examples~$\{\rvx_i \}_{i=1}^N$, network optimizer $\mathcal{Q}$.}
    \Output{Estimated parameters $\phi= \{\enc,\dec,\prior\}, \energy$.}
    \For{$t = 0:T-1$}{           
        \smallskip
            {\bf Primal-Step}:\\
        1. {\bf Mini-batch}: Sample observed examples $\{ \rvx_i \}_{i=1}^n$ \\
            2. {\bf Mini-batch}: Sample generated examples $\{ \hat{\rvx}_i \}_{i=1}^n$, where $ \hat{\rvx}_i\sim p_{\dec,\prior}(\hat{\rvx})$ \\
        3. {\bf Sample $\Tilde{\rvx}$ by MCMC}: For each generated $\hat{\rvx}_i$, sample $\Tilde{\rvx}_i$ using \cref{eq: LD_cond} for $K$ steps \\
        4. {\bf Learning $\energyfn$}: $\energy_{t+1} = \mathcal{Q}(\nabla_{\energy_{t}} \hat{\gL}(\energy_{t}),\energy_{t})$\\
        5. {\bf Learning and Calibrating VAE $\phi_t$}: $\phi_{t+1} = \mathcal{Q}( \nabla_{\phi_{t}}\hat{\gL}(\phi_{t}) + \lambda\nabla_{\phi_{t}} \hat{\gL}_{\mathrm{con}}(\phi_{t}),\phi_{t})$\\
            {\bf Dual-Step}:\\
            6. {\bf Update  $\lambda$}: update $\lambda$ according to \cref{eq:dual step}\\
        }
    \caption{\name Algorithm.}
    \label{algorithm:Primal-Dual}
\end{algorithm}

\section{More Details in Zero-Shot Image Restoration}
\label{app:detailed linear degraded operator}
Typical image restoration tasks usually have simple $\mA$ and $\mA^\dagger$, we give some examples following:

\spara{Colorization} $\mA = [1/3, 1/3, 1/3]$  converts each RGB channel pixel $[r,g,b]^T$ into a grayscale value $[r/3+g/3+g/3]$. A simple pseudo-inverse $\mA^\dagger$ is $\mA^\dagger = [1,1,1]^T$, that satisies $\mA\mA^\dagger\mA = \mA$.

\spara{Super Resolution} For SR with scale $n$, we can set $\mA\in\R^{1\times n^2}$ as a average-pooling operator $[1/n^2, \cdots, 1/n^2]$. A simple pseudo-inverse $\mA^\dagger\in\R^{n^2\times1}$ is $\mA^\dagger = [1,\cdots,1]^T$.

\spara{Inpainting} For $\mA$ is a mask operator, simply let $\mA^\dagger = \mA$, we can have $\mA\mA^\dagger\mA = \mA$.

\section{Experiment Setting Details in Image Generation}
\label{app:image_geneartion}
\spara{Evaluation Metric} We employ the FID score as the metric in most experiments. We compute the FID score between 50k generated samples and training images. On CelebA HQ we compute 30k generated samples and training images due to the dataset only containing 30k images.

\spara{Dataset Details} For STL-10,  we follow the procedure in AutoGAN and DC-VAE, and resize the STL-10 images to $32\times 32$. 

\spara{Ablations in \cref{tab:main_result}} The VAE and Flow use the same architecture as \shortname and EC-Flow, except the EBM is not needed in VAE and Flow. 

\spara{Energy-Calibrated Variational Learning} We use five MCMC steps for calibrating the posterior. We employ exactly the same architecture of VAE and EBM (only need for \shortname and \shortname w/ calibrated posterior) for all variants. Both the MSE and ELBO are measured on the test set of CIFAR-10.

\spara{Hyper-parameters} Given a large number of datasets, heavy compute requirements, and limited computational resources, we don't optimize the hyper-parameters carefully. Even if the backbone is not carefully selected or designed, it's expected that we can use other well-designed backbones to easily get better results. Also, we just roughly choose the learning rate from 1e-4 or 2e-4 in all experiments. Following most previous work~\cite{du2019implicit,clel,coopflow}, we utilize the Langevin Dynamics with a low temperature (e.g., 1e-5) and select a small step size from 1e-6 or 1e-5.
On all datasets, we train \shortname using the Adam optimizer. 
For a fair comparison, EMA is applied for the VAE component for all variants in our ablations.
For all other training details (e.g., detailed model architecture), we refer readers to our full code, which will be released after published. 

\section{Experiment details in Zero-Shot Image Restoration}
\label{app:image_restoration}
We use the well-trained \textbf{\name} in zero-shot image restoration by the proposed application method in \cref{sec:image restoration method} with 50 MCMC steps. 

\begin{figure}[!t]
    \centering
    \subfloat[CoopFlow without MCMC (FID=79.45)]{\includegraphics[width=0.46\linewidth]{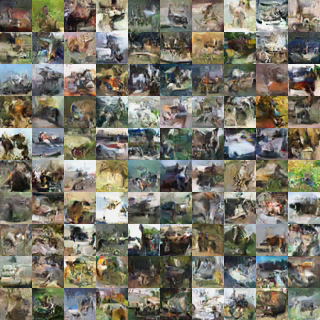}\label{fig:coopflow_without_mcmc}}
    \hfill
    \subfloat[VAEBM without MCMC (i.e., NVAE, FID=50.97)]{\includegraphics[width=0.46\linewidth]{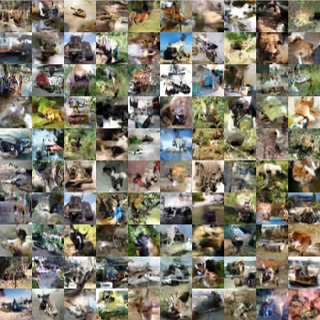}}
    \\
    \vspace{1em}
    \subfloat[CoopFlow w/o extra tuning hyper-parameters (FID=26.54)]{\includegraphics[width=0.46\linewidth]{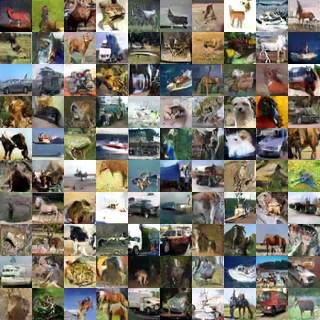}\label{fig:coopflow_untuned}}
    \hfill
    \subfloat[CoopFlow w/ carefully tuned hyper-parameters (FID=15.80)]{\includegraphics[width=0.46\linewidth]{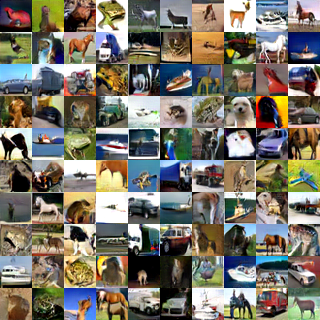}\label{fig:coopflow_tuned}}
    \\
    \vspace{1em}
    \subfloat[Real Data]{\includegraphics[width=0.46\linewidth]{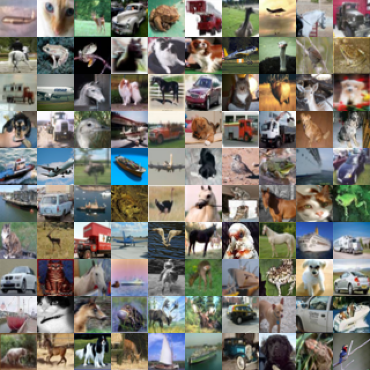}\label{fig:real_CIFAR}}
    \hfill
    \subfloat[\textbf{\shortname} (FID=5.20)]{\includegraphics[width=0.46\linewidth]{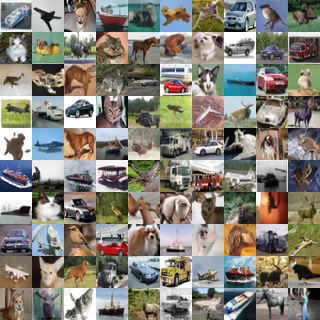}}\label{fig:EF_CIFAR100}
    \caption{Qualitative comparison of \name (Ours) and other models on CIFAR-10. Samples are un-selected.}
    \label{fig:comparison_CIFAR}
\end{figure}

\section{Experiment Details in Ablation Study}
\label{app:ablation}
\spara{Setting Details} The same architectural design of VAE's encoder, decoder, and energy function is consistently applied across all variants of \shortname in our ablation studies. For VAE+GAN, the same architectural design of VAE is applied, and we directly use the architecture of energy function as the discriminator. For EBM the same architectural design of energy function is applied. For EBM nit w/ VAE samples, the same architectural design of all networks is applied.
For \shortname w/ flow prior, we use a glow parameterized by MLPs as the architecture. For \shortname w/ LPIPS, we only use LPIPS in modeling $p_\dec(\rvx|\rvz)$, as we found the benefit of using LPIPS in modeling calibration loss is negligible.
We employed a replay buffer size of 10,000 for training the EBM, with a distribution of 5\% sampling from noise and 95\% sampling from the replay buffer. Regarding the EBM initialized with VAE samples, a similar approach was adopted, utilizing the same replay buffer size of 10,000, 5\% sampling from the VAE, and 95\% sampling from the replay buffer. Notably, in the absence of a replay buffer during the training of the EBM initialized with VAE samples, we were unable to produce reasonable generation outcomes. 

\subsection{Additional Ablations} 
\spara{Comparison to CoopVAEBM} We provide an extra comparison with CoopVAEBM~\cite{coopvaebm} which is a cooperative learning approach for training VAE and EBM. As discussed in related work section (\cref{sec:related_work}), the existing cooperative learning approach trains the base generative model (i.e., VAE) solely using generated samples, leading to biased learning. 
To further emphasize the necessity of training the base generative model (i.e., VAE) upon both real data and generated samples with adaptive weight as proposed in our \textbf{\shortname}, we train the CoopVAEBM using the same architecture as \shortname to ensure a fair comparison instead of directly using the reported results in their original paper which might be influenced by architectural differences. As shown in \cref{tab:addtional_ablation}, our \shortname outperforms CoopVAEBM by a large margin in terms of both FID, MSE, and ELBO. Additionally, it is observed that the MCMC still performs an important role in improving the generation quality of CoopVAEBM. 
These results indicate that our proposed \shortname is more effective with more accurate likelihood learning and better generation quality.

\spara{Effect of Exponential Moving Average (EMA)} The EMA technique  has been extensively incorporated into prior generative models~\cite{karras2020training,song2023consistency,song2020improved}, which demonstrates its widespread applicability and effectiveness. In alignment with these findings, we found the EMA can also enhance the performance of our model, \shortname, as indicated by the improved FID scores. As detailed in \cref{tab:addtional_ablation_new}, the application of EMA to \shortname results in a noteworthy reduction in the FID score from 8.02 to 5.20 on CIFAR-10.

\spara{Effect of Conditional Energy-Based Models} Unlike the previous cooperate approach~\cite{xie2018cooperative,coopvaebm,dualmcmc} which utilize unconditional EBMs for obtaining MCMC-revised samples, we utilize conditional EBMs to produce calibrated samples. Conditional EBMs have the advantage of constraining the high-density regions to be localized around the condition $\mathbf{x}$. This not only simplifies the calibrating process for the VAE but also enables us to focus on maximizing the conditional likelihood of the calibrated samples given condition samples via the VAE's decoder. Otherwise, an additional inference step would be necessary to infer the related latent variables of the MCMC-revised samples, as discussed in CoopVAEBM~\cite{coopvaebm}. Empirical evidence, as presented in \cref{tab:addtional_ablation_new}, supports the efficacy of conditional EBMs; the FID deteriorates from 5.20 to 5.89 when an unconditional EBM is employed, highlighting the superiority of the conditional approach.

\begin{table}[!tpb]
    \centering
    \small
    \footnotesize
\caption{Additional Quantitative results on CIFAR-10.}
\vspace{-2mm}
\label{tab:addtional_ablation}
\scalebox{0.85}{
   \begin{tabular}{lccc}
   \toprule
      \textbf{Model} & \textbf{FID$\downarrow$ }& \textbf{MSE$\downarrow$} & \textbf{ELBO$\uparrow$} \\
      \midrule
      CoopVAEBM w/ MCMC~\cite{coopvaebm} & 22.08 & - & -\\ 
      CoopVAEBM w/o MCMC~\cite{coopvaebm} & 26.17 & 0.0276 & 130.35\\ 
      \midrule
      \textbf{\shortname (ours)} & \textbf{5.20} & \textbf{0.0193} & \textbf{652.59}\\ 
\bottomrule
\end{tabular}
}
\end{table}

\begin{table}[!tpb]
    \centering
    \small
    \footnotesize
\caption{Additional Quantitative results on CIFAR-10.}
\vspace{-2mm}
\label{tab:addtional_ablation_new}
\scalebox{0.85}{
   \begin{tabular}{lccc}
   \toprule
      \textbf{Model} & \textbf{FID$\downarrow$ } \\
      \midrule
      \textbf{\shortname} & \textbf{5.20} \\ 
      \midrule
      \shortname w/o EMA & 8.02 \\ 
      \shortname w/ unconditional EBM & 5.89 \\ 
\bottomrule
\end{tabular}
}
\end{table}

\section{Additional Qualitative Results}
We present additional qualitative results in this section. Note that all images in this section are \textbf{un-selected}. 

\label{app:qualitative_results}
\spara{Qualitative Comparison to Other models} We provide a qualitative comparison to other models on CIFAR-10 in \cref{fig:comparison_CIFAR}. We use the official checkpoint from CoopFlow to generate its samples and use its official code to compute the corresponding FID score. For VAEBM w/o MCMC, we take the samples from its paper's appendix.

Please see if drop MCMC steps at test time sampling, how better our proposed method is compared to previous cooperative learning model~(i.e., CoopFlow) and previous work that combines generative model and EBMs~(i.e., VAEBM). And we note that CoopFlow is the most advanced model among existing cooperative learning models. Also, many existing works~(e.g., CoopFlow and VAEBM) related to EBMs need extra hyper-parameters tuning stage to achieve high performance or will lead to a significantly worse result. And it can be seen in \cref{fig:coopflow_tuned} that after carefully tuning the hyper-parameters, although the FID is better than the untuned, the color of samples by CoopFlow tends to be over-saturated,  which is not the property of real data. In contrast, our proposed \name can even drop MCMC steps at test time sampling while achieving high-quality samples.

\begin{figure*}[h]
    \centering
    \includegraphics[width = 0.9\linewidth]{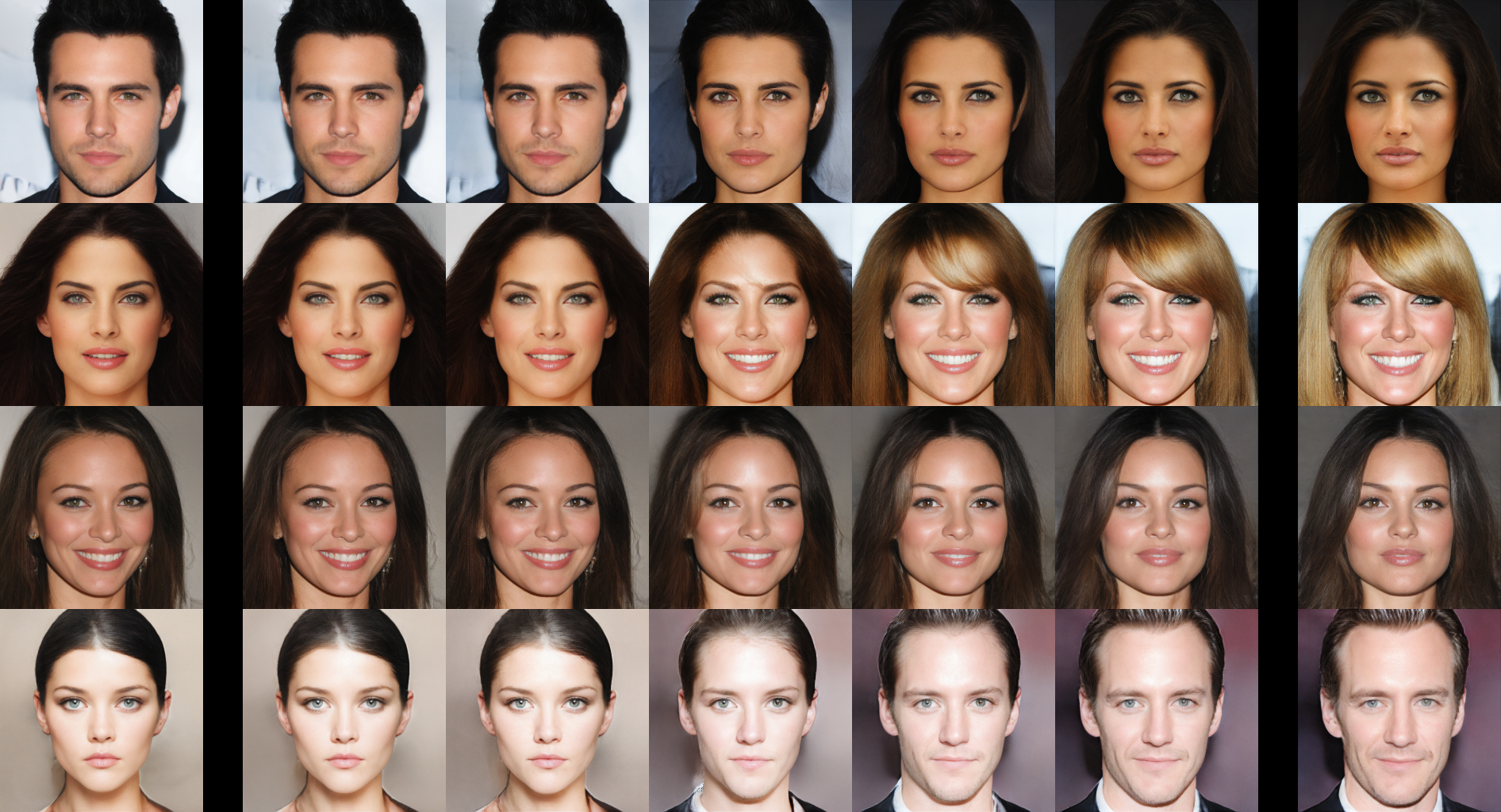}
    \caption{Interpolation results in latent space on CelebA-HQ-256.}
    \label{Interpolation images}
\end{figure*}

\begin{figure*}[h]
    \centering
    \includegraphics[width = 0.9\linewidth]{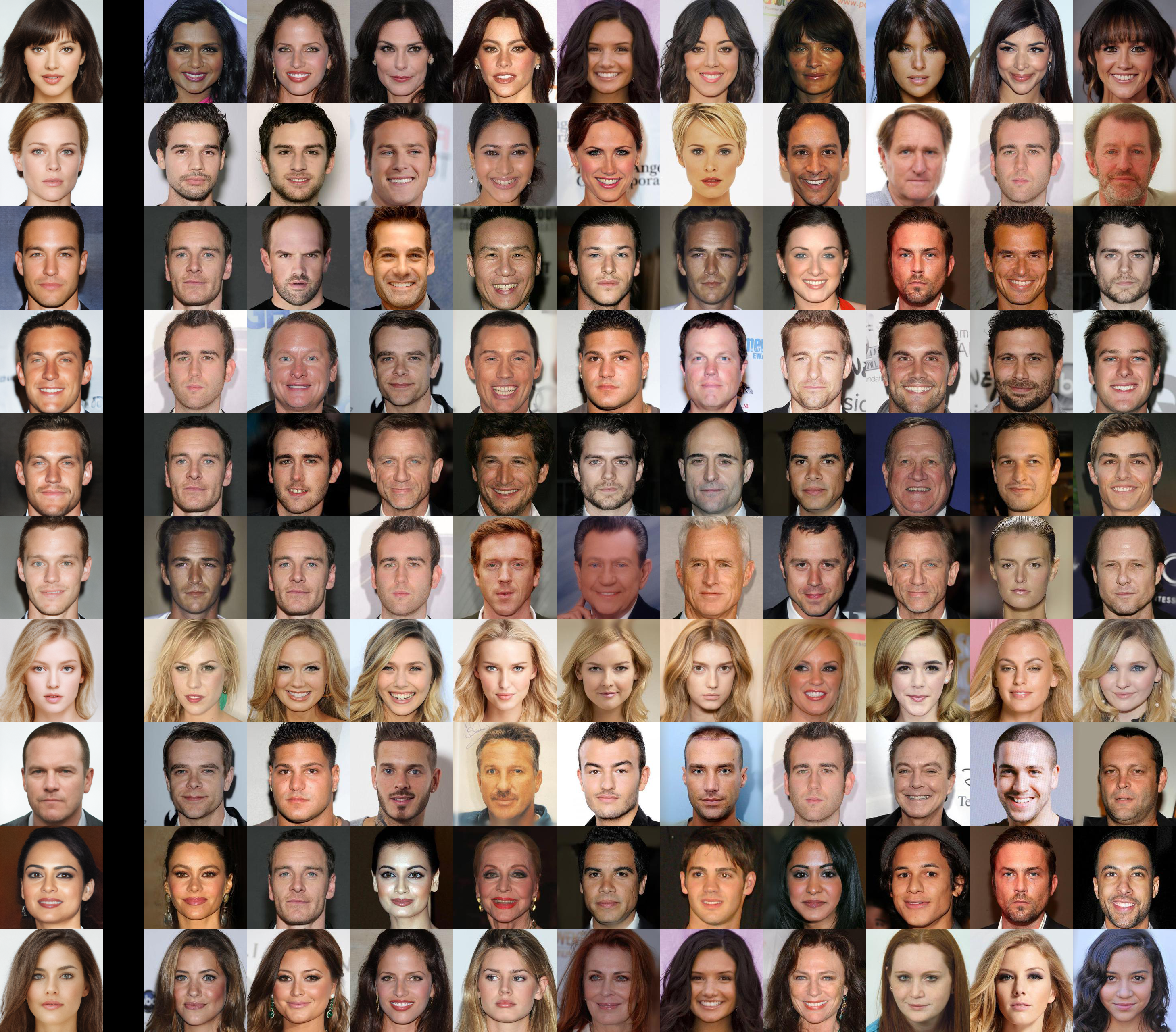}
    \caption{Generated images (left) and their nearest neighbors in VGG's feature space from the CelebA-HQ-256 training dataset.}
    \label{near celebahq}
\end{figure*}

\spara{Qualitative Results}
From the interpolation results in \cref{Interpolation images}, we conclude that our model has a good, smooth latent space.
We show the nearest neighbors from the training dataset with our generated samples in \cref{near celebahq}. It is easy to see our generated images are significantly different from images from the training dataset, which concludes that our model does not simply remember the data.

\begin{figure*}[h]
    \centering
    \includegraphics[width = 0.9\linewidth]{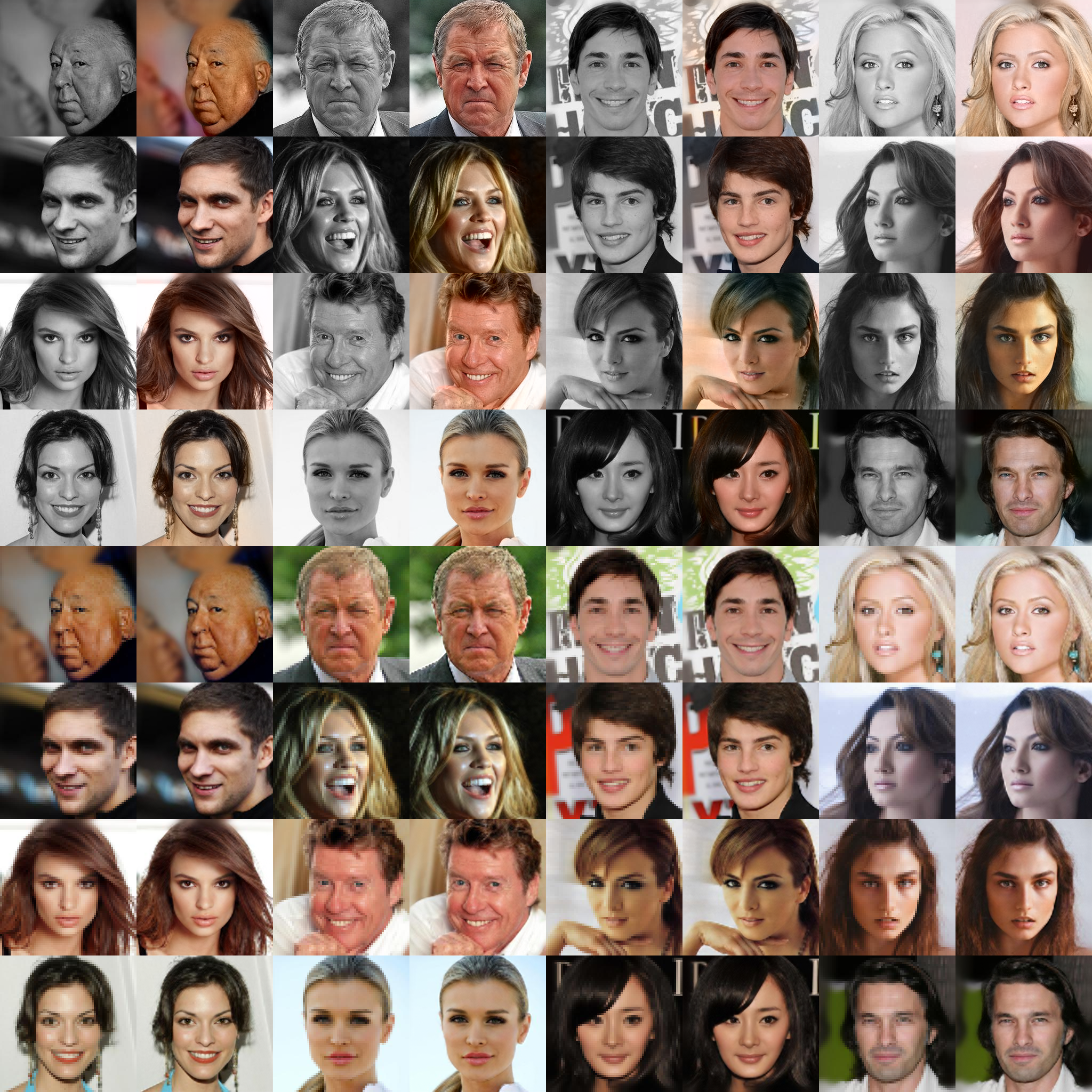}
    \caption{Additional Zero-Shot Image Restoration~(Colorization, 4$\times$ SR) Samples on CelebA-HQ-256. Samples are uncurated.}
    \label{Additional ir_celeba256}
\end{figure*}

Additional Zero-Shot Image Restoration samples are in \cref{Additional ir_celeba256}.

\begin{figure*}[h]
    \centering
    \includegraphics[width = 0.58\linewidth]{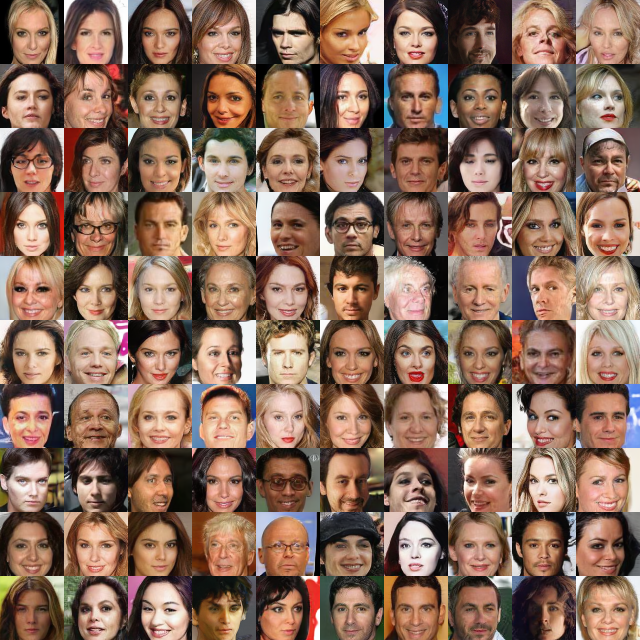}
    \caption{Additional samples from CelebA 64. Samples are uncurated.}
    \label{Additional celeba}
\end{figure*}

Additional samples on CelebA 64 are in \cref{Additional celeba}.

\begin{figure*}[h]
    \centering
    \includegraphics[width = 0.58\linewidth]{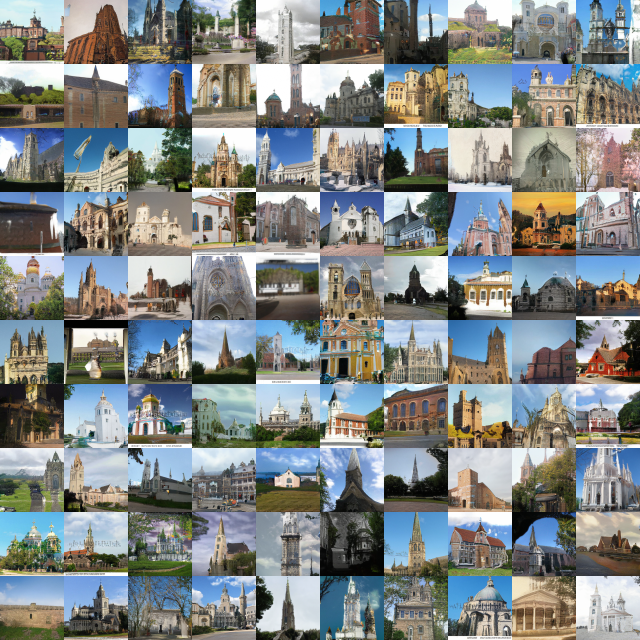}
    \caption{Additional samples from LSUN Church 64. Samples are uncurated.}
    \label{Additional church}
\end{figure*}

Additional samples on LSUN Church 64 are in \cref{Additional church}.

\begin{figure*}[h]
    \centering
    \includegraphics[width = 0.5\linewidth]{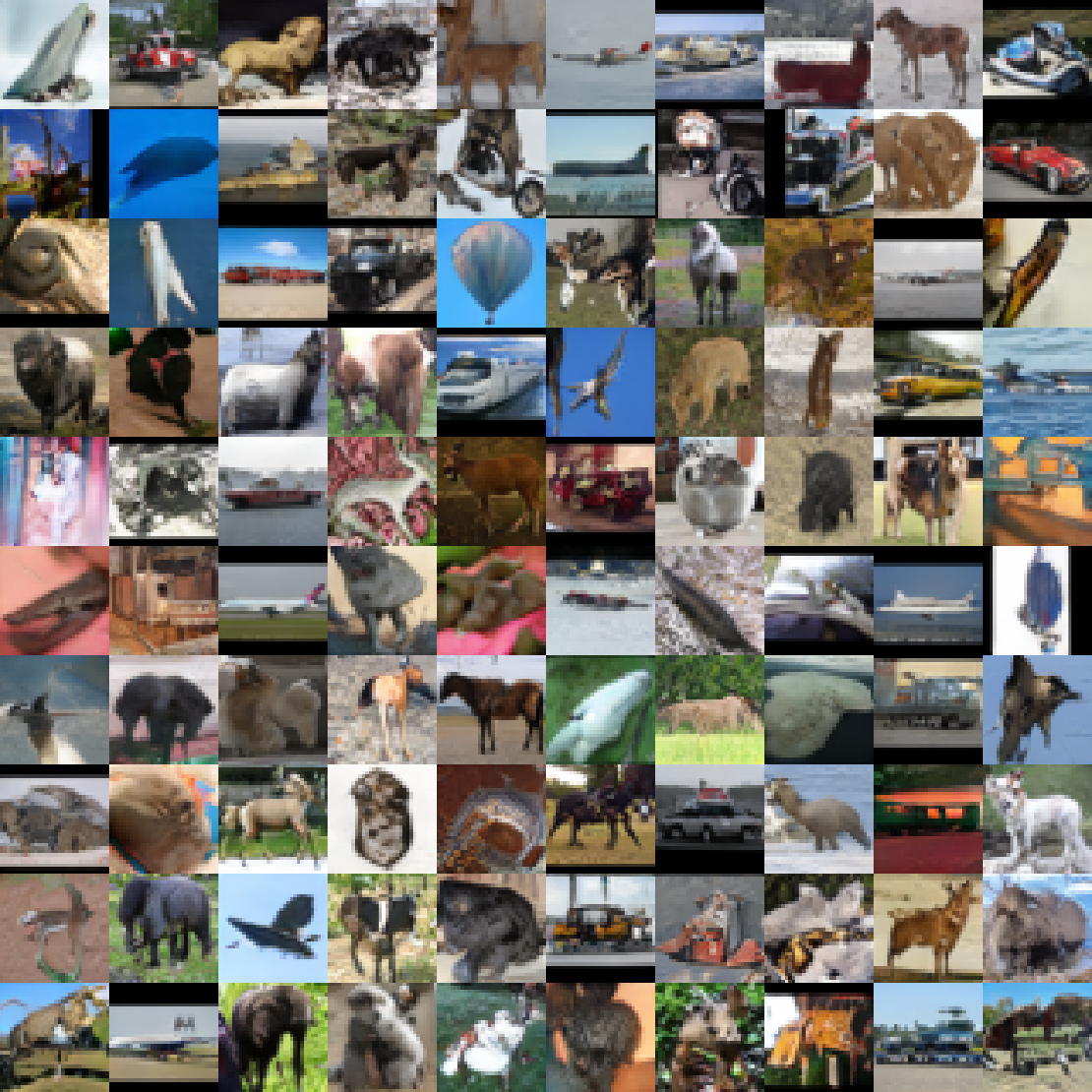}
    \caption{Additional samples from STL-10 which are uncurated.}
    \label{Additional STL}
\end{figure*}

Additional samples on STL-10 are in \cref{Additional STL}.

\begin{figure*}[h]
    \centering
    \includegraphics[width = 0.5\linewidth]{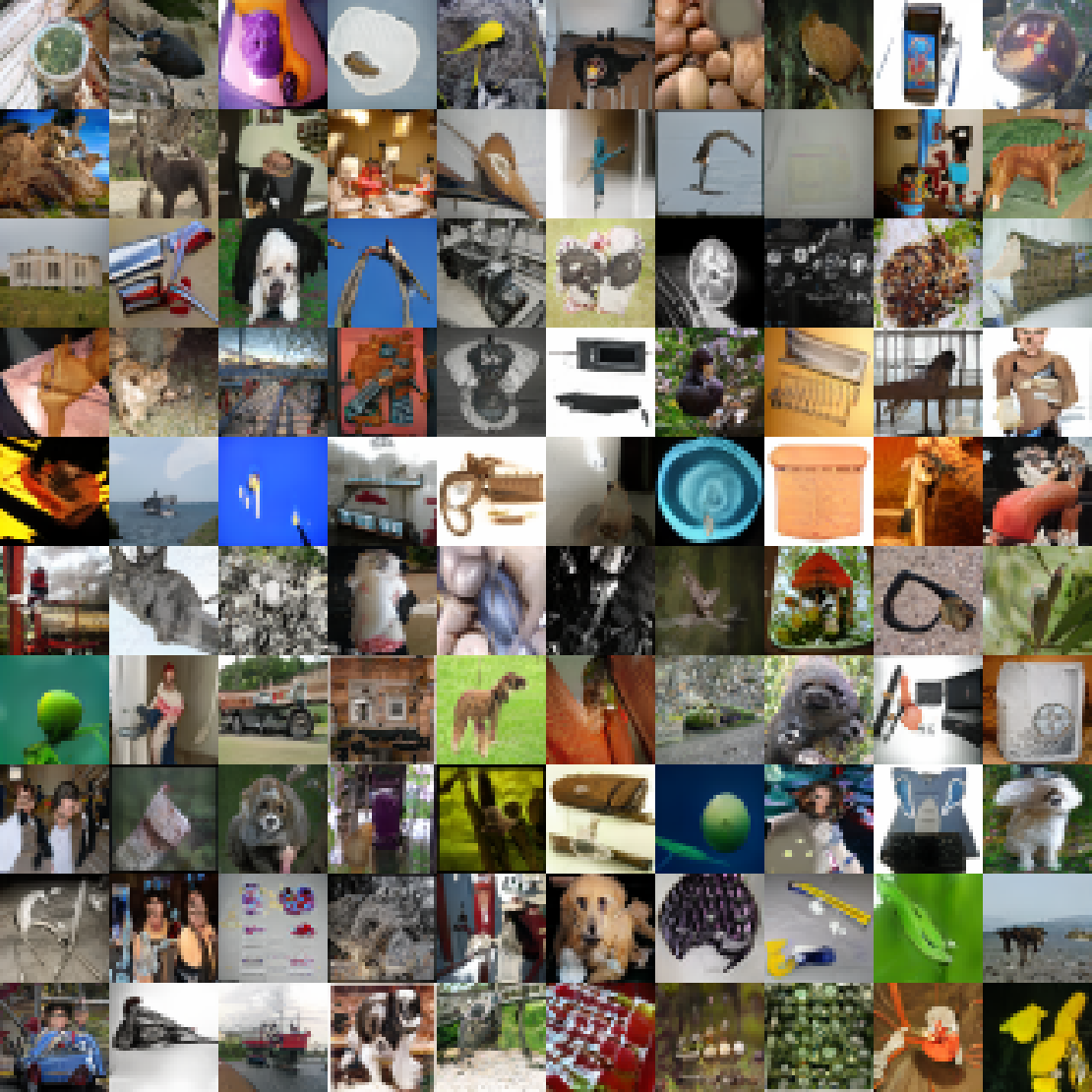}
    \caption{Additional samples from ImageNet 32 which are uncurated.}
    \label{Additional ImageNet}
\end{figure*}

Additional samples on ImageNet 32 are in \cref{Additional ImageNet}.

\begin{figure*}[h]
    \centering
    \includegraphics[width = 0.9\linewidth]{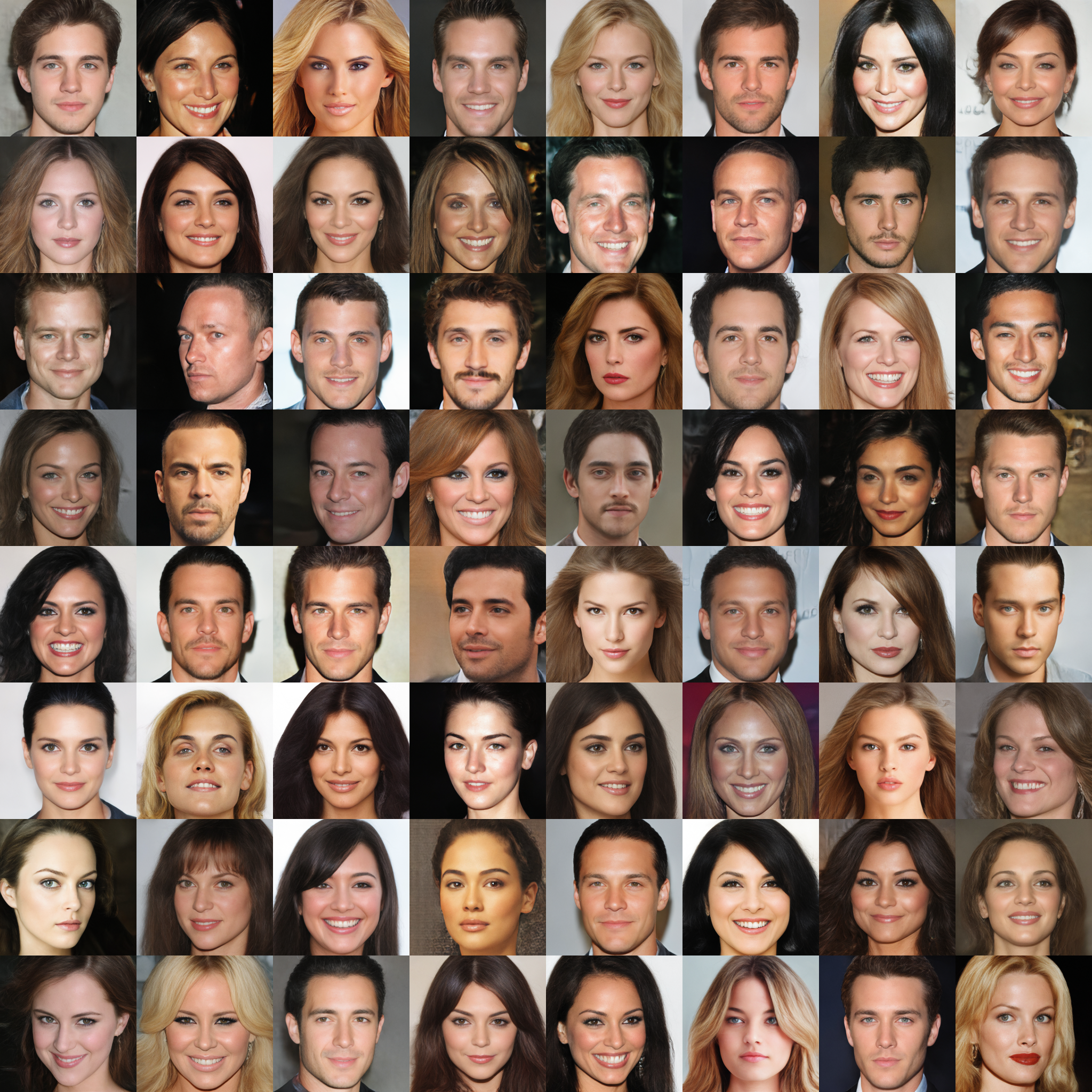}
    \caption{Additional samples from CelebA-HQ-256. Samples are uncurated.}
    \label{Additional celebahq}
\end{figure*}

Additional samples on CelebA-HQ-256 are in \cref{Additional celebahq}.

Interpolation results on CelebA-HQ-256 are in \cref{Interpolation images}.

Samples on CelebA-HQ-256 and their nearest neighbors from the training dataset are in \cref{near celebahq}.

\end{document}